\documentclass[10pt,journal,compsoc]{IEEEtran}
\ifCLASSOPTIONcompsoc
  \usepackage[nocompress]{cite}
\else
  \usepackage{cite}
\fi
\usepackage{amsmath,graphicx}
\usepackage{bm}
\usepackage{caption}
\usepackage{hyperref}
\usepackage{amsmath}
\usepackage{amsfonts}
\usepackage{amssymb}
\usepackage{color}
\usepackage{subfigure}
\usepackage{booktabs}
\usepackage{arydshln}
\usepackage{makecell}
\usepackage{colortbl}
\usepackage{multirow}
\usepackage{multicol}
\usepackage{ragged2e}
\usepackage{textcomp,booktabs,xcolor}
\ifCLASSINFOpdf
\else
\fi
\begin{document}
%
\title{Meta-Learn Unimodal Signals with Weak Supervision for Multimodal Sentiment Analysis}

\author{\IEEEauthorblockN{Sijie Mai\IEEEauthorrefmark{1}$^{,1}$,
		Yu Zhao\IEEEauthorrefmark{2}$^{,1}$,
		Ying Zeng\IEEEauthorrefmark{1},
		Jianhua Yao\IEEEauthorrefmark{2},
Haifeng Hu\IEEEauthorrefmark{1}}\\
\IEEEauthorblockA{\IEEEauthorrefmark{1}School of Electronics and Information Technology, Sun Yat-sen University, Guangzhou, China\\}
\IEEEauthorblockA{\IEEEauthorrefmark{2}Tencent AI Lab, Tencent, Gemdale Viseen Tower B, No. 16, Gaoxin South 10th Road, Shenzhen, China}
\IEEEcompsocitemizethanks{\IEEEcompsocthanksitem Haifeng Hu and Jianhua Yao are corresponding authors.\protect\\
	E-mails: huhaif@mail.sysu.edu.cn, jianhua.yao@gmail.com\protect
	\IEEEcompsocthanksitem  Sijie Mai and Yu Zhao contribute equally to this work.}
}

\IEEEtitleabstractindextext{%
\begin{abstract}
Multimodal sentiment analysis aims to effectively integrate information from various sources to infer sentiment, where in many cases there are no annotations for unimodal labels. Therefore, most works rely on multimodal labels for training. However, there exists the noisy label problem for the learning of unimodal signals as multimodal annotations are not always the ideal substitutes for the unimodal ones, failing to achieve finer optimization for individual modalities. In this paper, we explore the learning of unimodal labels under the weak supervision from the annotated multimodal labels. Specifically, we propose a novel meta uni-label generation (MUG) framework to address the above problem, which leverages the available multimodal labels to learn the corresponding unimodal labels by the meta uni-label correction network (MUCN). We first design a contrastive-based projection module to bridge the gap between unimodal and multimodal representations, so as to use multimodal annotations to guide the learning of MUCN. Afterwards, we propose unimodal and multimodal denoising tasks to train MUCN with explicit supervision via a bi-level optimization strategy. We then jointly train unimodal and multimodal learning tasks to extract discriminative unimodal features for multimodal inference. Experimental results suggest that MUG outperforms competitive baselines and can learn accurate unimodal labels.
\end{abstract}

\begin{IEEEkeywords}
Multimodal Sentiment Analysis, Meta-Learning, Unimodal Label Learning, Weakly Supervised Learning
\end{IEEEkeywords}}

\maketitle

\IEEEdisplaynontitleabstractindextext

%
\IEEEpeerreviewmaketitle

\section{Introduction}\label{sec:introduction}
\IEEEPARstart{W}{ith}  the swift advancement of technology and the proliferation of social media platforms, multimodal data have become increasingly prevalent for the execution of a multitude of downstream applications, including multimodal language analysis \cite{MAG-BERT}, multi-omics integrative analysis \cite{mvib}, human action recognition \cite{human_action}, etc.  
Multimodal sentiment analysis (MSA) \cite{Poria2017A,m3sa}, which seeks to extract human sentiments and viewpoints from language, acoustic, and visual data streams, has recently garnered considerable interest.
Researchers endeavor to devise effective models encompassing variations of recurrent neural networks (RNNs) \cite{HFFN,Zadeh2018Memory,Zadeh2018Multi,MFRM}, attention-based networks \cite{MRM,sun2023modality,chen2022weighted_attention_taslp,sun2022tensorformer,10572356}, BERT-based models \cite{MAG-BERT,li2023interpretable_LLM_MSA} and so on to capture sufficient interactions between modalities and learn joint embedding in a common manifold. 
Predominantly, these methodologies concentrate on learning rich multimodal representations to uncover sophisticated cross-modal dynamics, and they demonstrate promising results in MSA.

\begin{figure}[h]
	\centering
	\includegraphics[width=1.0\linewidth]{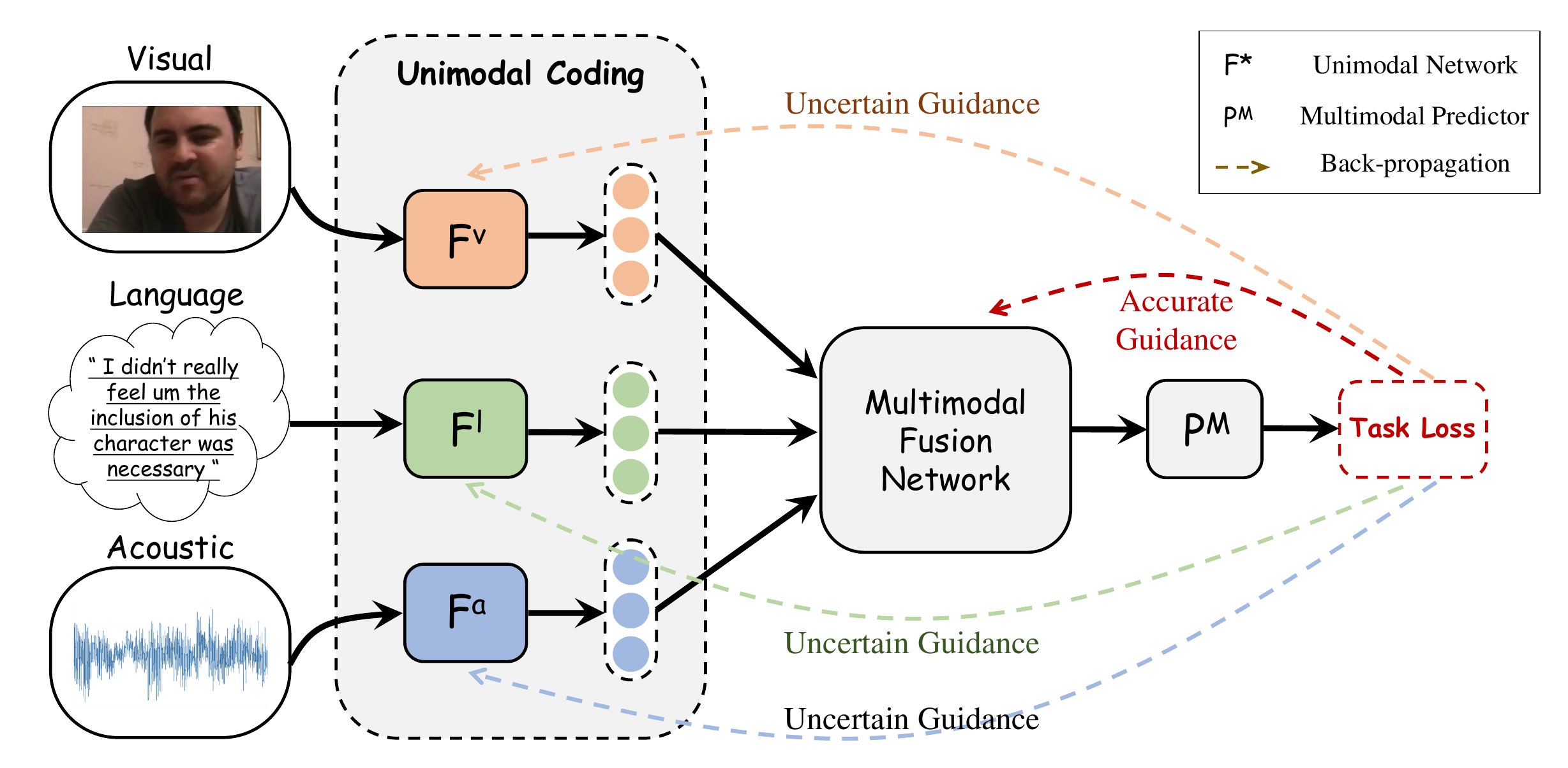}
	\caption{\label{motivation}\textbf{The noisy label problem for unimodal learning.} The multimodal label can be regarded as an uncertain guidance for each unimodal network.}
	\vspace{-0.2cm}
\end{figure}

However, due to the subtle and complex expression of human
beings, individual modalities may convey distinct aspects of sentiment, leading to the label inconsistency problem between multiple modalities and hindering the performance of the model \cite{wang2024evaluation,hirano2021recognizing}.
In typical multimodal representation learning tasks, annotations of unimodal signals are often unavailable. As shown in Fig.~\ref{motivation}, existing algorithms mostly rely on the multimodal learning loss to update the whole model, which might not be able to extract discriminative and meaningful unimodal features for better learning of downstream tasks. A few works try to define unimodal labels as the corresponding global label of the multimodal sample, and conduct various learning tasks based on the defined unimodal labels \cite{mai2022curriculum}. Nevertheless, the global multimodal label might not be a suitable substitute of unimodal labels  under some circumstances because unimodal representations fail to convey the global information of a multimodal sample and only reveal a specific perspective of the sample. Some prior works \cite{mmsa,SUGRM} try to calculate unimodal labels based on the distance between the corresponding unimodal representations and the positive/negative centres of multimodal representations in a non-parametric way. However, these methods are not learnable and have limited expressive power, and the quality of the calculated unimodal labels is hard to estimate.

In this paper, we present a novel multimodal learning framework to excavate the rich information stored in the unimodal signals for multimodal learning. The proposed framework aims to learn unimodal labels under the weak supervision from multimodal annotations, and then jointly trains unimodal and multimodal learning tasks to leverage the available information to the utmost extent. Specifically, we propose a novel meta-learning framework to leverage the available multimodal labels to help learn unimodal labels, which is more expressive and discriminative than using non-parametric approaches to calculate unimodal labels as in previous algorithms of MSA \cite{mmsa,SUGRM}. To the best of our knowledge, we are the first to introduce meta-learning \cite{9} into MSA for learning accurate unimodal labels. Our framework is divided into three stages. In the first stage, we train the whole network with the human-annotated multimodal labels, and design a contrastive-based projection module to maximize the mutual information and narrow down the gap between unimodal and multimodal representations. The unimodal predictors are also trained using the projected multimodal representations to provide discriminative power at this stage. In the second stage, a novel meta-learning strategy is proposed to use multimodal labels to guide the learning of unimodal labels for individual modalities, which can estimate the quality of learned unimodal labels and help to meta-update the meta uni-label correction network (MUCN) via a bi-level optimization strategy. Different from previous works on noisy label learning that merely rely on the loss of clean samples to meta-update the label correction network \cite{mlc,noisy_cvpr_2023,wu2021learning},  we elaborately design unimodal and multimodal denoising tasks to train MUCN in a successive manner  with explicit supervision signal, enabling the effective learning of MUCN. The proposed bi-level optimization strategy does not involve the update of other modules (e.g., the main net) as in other noisy label learning algorithms based on meta-learning \cite{mlc,noisy_cvpr_2023,wu2021learning}, which is stable in training. In the final stage, we jointly train multimodal learning task and unimodal learning tasks using the learned unimodal labels to extract more expressive and discriminative unimodal features for further fusion.

In brief, the contributions of this paper can be summarized as:
\begin{itemize}
	\item We elaborately devise a three-stage multimodal framework, named meta uni-label generation (MUG), to learn unimodal labels and then jointly train unimodal and multimodal learning tasks. In this way, we can leverage the available multimodal information to the utmost extent and extract more discriminative and expressive unimodal features for the better learning of multimodal tasks.
	\item We propose a novel meta-learning algorithm to leverage multimodal labels for learning accurate unimodal labels. In the proposed meta-learning algorithm, we design the contrastive-based projection module to narrow down the gap between unimodal representations and the projected multimodal representations at the first stage.  In this way, we can use the available multimodal labels to estimate the quality of learned unimodal labels, and then meta-update MUCN. Furthermore, compared to other meta-learning algorithms \cite{mlc,wu2021learning,noisy_cvpr_2023}, we elaborately design unimodal and multimodal denoising tasks to train MUCN with explicit supervision at the meta-learning stage, which enables a more stable learning of MUCN. 
	\item The proposed MUG outperforms other algorithms on the task of MSA across three datasets. In addition, MUG outperforms other meta-learning based noisy label learning algorithm \cite{mlc}.
\end{itemize}


\section{\textbf{Related Work}}\label{sec:Related Work}
\label{sec:format}
\subsection{Multimodal Sentiment Analysis}
MSA has garnered considerable interest due to its ability to decipher human language and extract sentiments from spoken words, acoustic signals, and visual cues. A plethora of existing studies has concentrated on devising fusion techniques aimed at learning rich and informative multimodal embeddings \cite{Poria2017A,cubemlp_mm2022,MSA_survey_fusion}.
Specifically, methods that are based on tensor fusion \cite{T2FN,Zadeh2017Tensor,koromilas2023mmatr_tensor} have drawn considerable attention, as they are capable of learning a unified multimodal representation endowed with high expressiveness. 
To highlight important interactions, numerous scholars propose cross-modal attention mechanisms \cite{Zadeh2018Multi, sun2023modality,chen2022weighted_attention_taslp, taslp_attention, MULT}. 
Following the groundbreaking success of BERT \cite{BERT}, there has emerged a trend toward fine-tuning large pre-trained Transformer models using multimodal datasets \cite{MAG-BERT,li2023interpretable_LLM_MSA}. For instance, All-modalities-in-One BERT (AOBERT) \cite{aobert} introduces multimodal masked language modeling and alignment prediction tasks, which serve to further pretrain BERT, thereby deeply mining the correlations within and between modalities.
Moreover, certain methodologies employ tools such as Kullback–Leibler divergence (KL-divergence) and canonical correlation analysis to regularize the learning process of unimodal distributions, thereby mitigating the distributional disparities among different modalities \cite{ICCN,nips_cca,nll}.
For instance, Sun  \textit{et al.} \cite{sun_meta} apply KL-divergence and propose a meta-learning algorithm to learn favorable unimodal representations for multimodal tasks. But they still use multimodal loss to meta-learn unimodal representations, whose objective and algorithm are totally different from our meta-learning strategy.

Recently, self-supervised and weakly-supervised learning on multimodal data has attracted significant attention \cite{vision_language_vilbert,robust_contrastive_multi_view,goel2022cyclip,COBRA,boosting}. 
For instance, Chen \textit{et al.} \cite{chen2017predicting} propose a weakly-supervised algorithm to learn convolutional neural networks using cheaply available emotion labels that contain noise. They use a probabilistic graphical model to simultaneously learn discriminative multimodal descriptors and infer the confidence of label noise, such that the label noise can be filtered out. Different from them, we focus on learning unavailable unimodal labels in a weakly-supervised manner. Furthermore,
modality correlation learning, which uses a predictor to predict the correlation score of two or multiple modalities, is an effective strategy of weakly-supervised learning \cite{mcl,mai2022curriculum}. However, this line of research simply defines each unimodal label as the corresponding multimodal label to conduct weakly-supervised learning, while our method leverages weakly-supervised learning to learn accurate unimodal labels based on the available multimodal labels.
Another popular strategy of self-/weakly-supervised learning is to apply contrastive learning to learn representations from multimodal data  \cite{COBRA,contrastive_comir,robust_contrastive_multi_view,goel2022cyclip,confede,li2024cormult}.
For example,
Hybrid Contrastive learning (HyCon) \cite{hycon} introduces intra-modal and inter-modal contrastive learning strategies that pushes unimodal representations from the same category closer and pushes those from different categories apart, and Contrastive FEature DEcomposition (ConFEDE) \cite{confede} simultaneously performs contrastive representation learning and feature decomposition of unimodal representations to enhance the generated multimodal representations. Moreover, Anand \textit{et al.} \cite{anand2023multi} propose a label consistency calibration loss to prevent label bias by enabling calibration of the peer-ensembled fusion branches with respect to the difficulty of different emotion labels, and design a multimodal distillation loss to calibrate the fusion network by minimizing the KL-divergence with the modality-specific peer networks. They also conduct self-supervised unimodal-unimodal and unimodal-multimodal contrastive learning to improve the generalization of the model across diverse speaker backgrounds. 
In contrast, we perform unimodal-multimodal contrastive learning for a different purpose, aiming to narrow down the distribution gap between unimodal features and the projected multimodal representations.

Although achieving promising performance, these methods do not consider the learning of unimodal labels, leading to the sub-optimal performance of the model due to the insufficient learning of unimodal features. Different from them, we propose a novel meta-learning strategy to learn unimodal labels based on the available multimodal labels, and conduct multi-task training of unimodal and multimodal learning tasks to extract favorable unimodal features for a more robust multimodal framework.

\subsection{Label Ambiguity and Inconsistency}
In MSA, the ground-truth labels given by multiple annotators often disagree for the reason that the annotation of sentiments and emotions is a subjective and ambiguous task \cite{hirano2021recognizing,yannakakis2018ordinal}. Therefore, the annotated labels of some samples might be inaccurate, which is referred to as the label ambiguity problem. Yannakakis \textit{et al.} \cite{yannakakis2018ordinal} reveal that assigning absolute values to emotions and sentiments is not only a noisy task, but also is unsuitable due to their subjective nature. They argue that  ordinal labels are more suitable to represent emotions and sentiments, suggesting that assigning relative values to subjective concepts is better aligned with the underlying representations than assigning absolute values. Based on this view, they design the preference learning paradigm for training affective models using ordinal data to verify the benefits of relative annotation in affective computing \cite{yannakakis2018ordinal}.
In addition, to deal with the situation that the annotated labels are not necessarily accurate, Hirano \textit{et al.} \cite{hirano2021recognizing} introduce a weakly-supervised learning algorithm called tri-teaching to identify and remove the noisy labels.
In contrast, Lotfian \textit{et al.} \cite{lotfian2019curriculum} resolve the label ambiguity issue from a different perspective, using curriculum learning to train the model where training samples are gradually presented in increasing level of difficulty. They assume that the ambiguous samples for humans are also ambiguous for machines, and the performance can be improved by prioritizing less ambiguous samples (i.e., easier samples) at the beginning of the training process of the deep neural network.  While these methods and our algorithm both aim to handle labelling issues, their focuses are different. Different from these methods that handle the label ambiguity issue of sentiments and emotions from various perspectives, our method mainly deals with the label inconsistency problem between different modalities within each multimodal sample. In other words, our MUG focuses on learning accurate unimodal labels for each modality instead of removing the label noise of multimodal samples. The handling of the label ambiguity problem in MSA is left for our future work. 

The works that address the label inconsistency problem between various modalities are closely related to our algorithm.
MSA infers sentiment from language, acoustic, and visual modalities. However, due to subtle and nuanced expression of human beings, different modalities may convey distinct aspects of sentiment, leading to the inconsistency between various modalities and hindering the performance of the model \cite{wang2024evaluation,SUGRM}. Wang \textit{et al.} \cite{wang2024evaluation} reveal that significant performance degradation of traditional models and multimodal large language models occurs when semantically conflicting multimodal data are selected for evaluation. 
When handling the missing modality problem, Zeng \textit{et al.} \cite{zeng2022mitigating} also discover the label inconsistency issue between modalities where the sentiment of a sample may change when a key modality is absent, resulting in the inaccurate prediction results. To resolve this issue, they propose an ensemble-based missing modality reconstruction network to recover semantic features of the key missing modality, and then check the semantic consistency to determine whether the absent modality is crucial to the overall sentiment polarity \cite{zeng2022mitigating}. Compared to these works, our algorithm addresses the label inconsistency issue from a different perspective. Instead of avoiding the negative effect of label inconsistency, we aim to learn accurate unimodal labels for individual modalities, which can extract more discriminative and expressive unimodal features for multimodal inference.


\subsection{Unimodal Label Learning}
Different modalities belonging to the same multimodal sample might express distinct sentiments.
In typical multimodal representation learning tasks, annotations of unimodal signals are often unavailable. 
To obtain accurate unimodal labels for learning more favorable unimodal representations,
Self-MM \cite{mmsa} proposes a self-supervised non-parametric approach to define unimodal labels and extract expressive unimodal features. It first calculates the positive/negative centres of multimodal representations according to multimodal annotations, and then defines unimodal labels based on the distance between the corresponding unimodal representations and the positive/negative centres.
Moreover, SUGRM \cite{SUGRM} improves Self-MM \cite{mmsa} by projecting features of each modality into a common semantic feature space, which allows simpler calculation of the unimodal labels and avoids unstable training. Different from these works, we propose a learnable meta-learning framework to generate unimodal labels in a weakly-supervised way, which is more expressive and can estimate the quality of learned unimodal labels. The proposed framework trains MUCN in an effective and stable way via the elaborately designed unimodal and multimodal denoising tasks, enabling MUCN to generate high-quality unimodal labels. To our best knowledge, we are the first to develop meta-learning based unimodal label learning algorithm in MSA.

The learning of unimodal labels can be regraded as a noisy label learning problem. Actually,
the learning of noisy labels has been developed in various research areas and shows promising results \cite{mlc,algan2021meta,xia2023tcc,noisy_cvpr_2023,sukhbaatar2015training,wu2021learning}. For example, Zheng \textit{et al.} \cite{mlc} propose a classic meta-learning method to learn the label of noisy samples given clean samples. However, the training of its label correction network (LCN) does not have explicit supervision signals but merely relies on the main task loss of clean samples. Moreover, it needs to first update the main net using the corrected labels generated by LCN  and then computes the meta-gradients of LCN, where the long bi-level back-propagation might weaken the effect of meta-gradients for updating LCN and lead to gradient vanishing. The simultaneous optimization over both main net and label correction network is difficult to achieve an optimal routine, limiting the representation ability of the network and accuracy of corrected labels \cite{noisy_cvpr_2023}. In contrast, we elaborately design unimodal and multimodal denoising tasks to learn MUCN with explicit supervision. Moreover, our bi-level optimization does not involve the update of other modules, which is more stable in training.

\section{\textbf{Algorithm}}\label{sec:Algorithm}
\begin{figure*}
	\centering
	\includegraphics[width=1.00\linewidth]{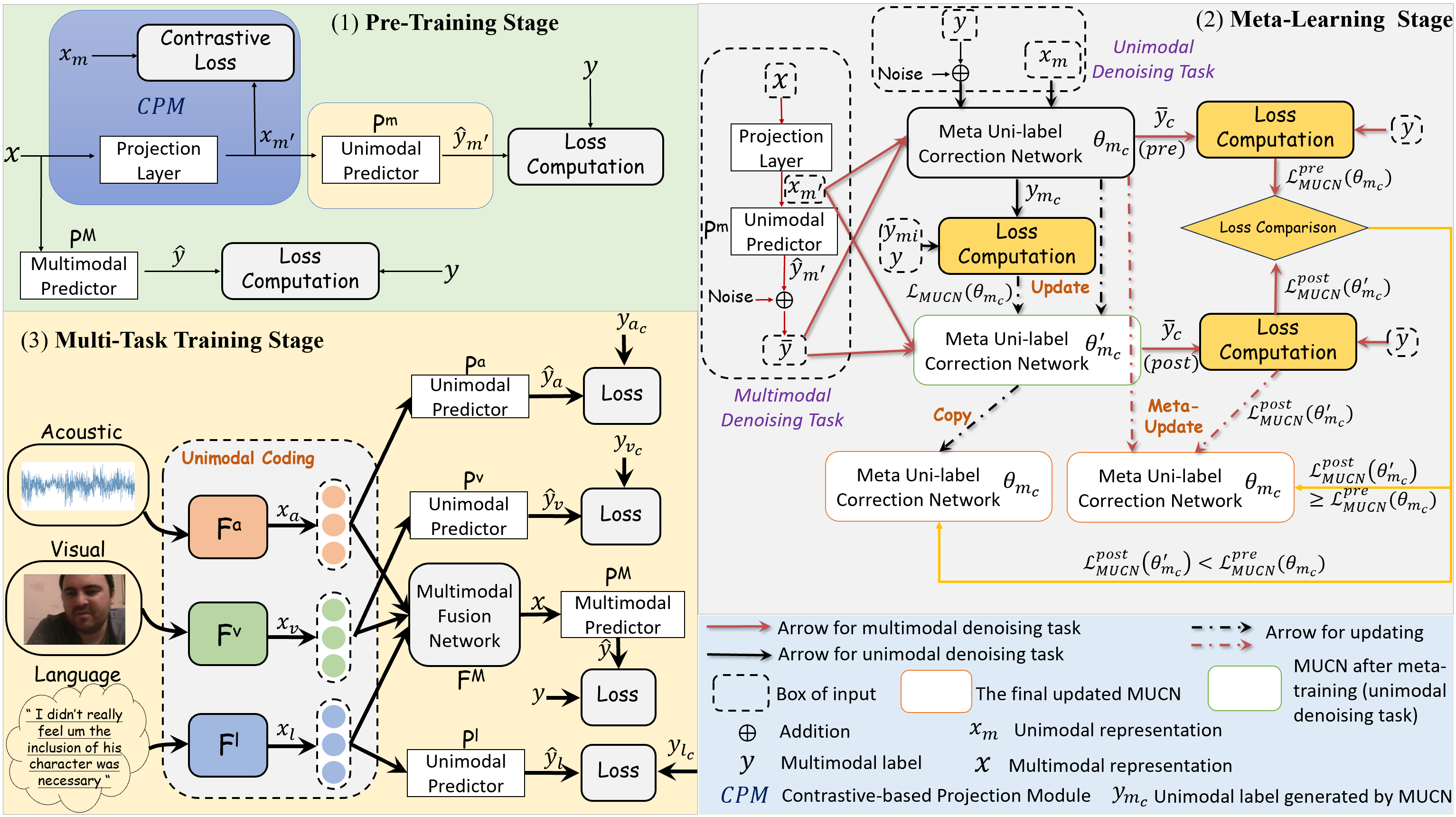}
	\caption{\label{main}\textbf{The diagram of the proposed MUG.} 
	}
\end{figure*}

In this section, we describe the proposed MUG in detail, which aims to learn accurate unimodal labels and then extract more discriminative unimodal features for multimodal fusion. MUG is divided into three stages, namely pre-training, meta-learning and joint training. The three stages and their interrelationships can be summarized as follows:
(1) Pre-training Stage: This initial phase serves two purposes. Firstly, it narrows down the distribution gap between unimodal representations and the projected multimodal representations through a contrastive-based projection module. Secondly, it initializes the parameters of the framework and cultivates efficient multimodal and unimodal representations. The primary objective here is to prepare the framework for subsequent meta-learning stage.
(2) Meta-learning Stage: In the second stage, we leverage the initialized framework and unimodal/multimodal representations to execute meta-learning of unimodal labels. This stage aims to yield proficient unimodal labels for the ensuing multi-task training stage.
(3) Multi-task Training Stage: The third stage involves two facets: unimodal learning tasks and multimodal learning task. Leveraging unimodal labels generated in the second stage, we conduct unimodal learning tasks. Simultaneously, the framework addresses multimodal learning task using the available multimodal labels. This conjoined effort refines unimodal and multimodal representations to enhance inferential outcomes. 
Importantly, the first two stages can be decoupled and trained independently, which significantly reduces the training time. By performing the first two stages only once, we attain the necessary unimodal labels for the joint training stage. The diagram of MUG is shown in Fig.~\ref{main}. 

Our downstream task is MSA, where the model is fed an utterance \cite{Olson1977From} (a segment of video demarcated by a sentence). Each utterance encompasses three modalities: acoustic (denoted as $a$), visual (denoted as $v$), and language (denoted as $l$). The objective of MSA is to infer a sentiment score based on the feature sequences of these three modalities. A conventional multimodal framework encompasses the extraction of unimodal representations, the amalgamation of these unimodal representations into a multimodal representation, and the execution of inference predicated on the integrated multimodal representation.
The summary of notations and acronyms is shown in Table~\ref{t1111}. In the following subsections, we illustrate the procedures of the proposed MUG in detail.

\begin{table*}[t]
	\centering
	\caption{ \label{t1111}\textbf{Table of Important Notations and Acronyms.}}
	\resizebox{1.8\columnwidth}{!}{\begin{tabular}{c|c}
			\hline
			Notations/Acronyms & Descriptions \\
			\hline
			MSA  & Multimodal Sentiment Analysis\\
			MUG  & Meta Uni-label Generation Framework\\
			MUCN  & Meta Uni-label Correction Network\\
			$MUCN_m$ & The meta uni-label correction network for modality $m$\\
			CPM & Contrastive-based Projection Module\\
			$\bm{x}_{m}$  &  The unimodal representation of modality $m$ (omitting the superscript) \\
			$\bm{x}_{m}^j$  &  The unimodal representation for modality $m$ of sample $j$ \\
			$\bm{x}_{m'}$  &  The projected multimodal representation of modality $m$ \\
			$\bm{x}$  & The multimodal representation\\
			$y$  & The multimodal label \\
			$\hat{y}$  & The multimodal prediction generated by multimodal predictor based on $\bm{x}$ \\
			$\hat{y}_{m'}$  & The prediction generated by unimodal predictor of modality $m$ based on $\bm{x}_{m'}$ \\
			$y_{mc}$  & The corrected label of modality $m$ generated by $MUCN_m$ \\
			$\Bar{y}$  & The corrupted noise label, input of multimodal denoising task \\
			$\Bar{y}_c$  & The corrected label of multimodal denoising task (generated by MUCN) \\
			$y_{mi}$  & The supervised signal of unimodal denoising task at later epochs of meta-learning \\
			$\hat{y}_{m}$  & The output of unimodal predictor based on unimodal representation $\bm{x}_{m}$ \\
			$\alpha$ & The meta-training learning rate\\
			$\alpha'$ & The learning rate of meta-updating\\
			$\theta_{m_c}$ & The parameter of meta uni-label correction network $MUCN_m$\\
			$\theta'_{m_c}$ & The updated parameter of $MUCN_m$ after unimodal denoising task\\
			$\bm{P}^m$ & The unimodal predictor of modality $m$\\
			$\bm{P}^M$ & The multimodal predictor that takes as input $\bm{x}$ and output multimodal prediction $\hat{y}$\\
			$\mathcal{L}_{MUCN}(\theta_{m_c})$ & The loss of unimodal denoising task of modality $m$\\
			$\mathcal{L}_{MUCN}^{pre}(\theta_{m_c})$ & The loss of multimodal denoising task of modality $m$ before performing unimodal denoising task and updating\\
			$\mathcal{L}_{MUCN}^{post}(\theta'_{m_c})$ & The loss of multimodal denoising task of modality $m$ after performing unimodal denoising task and updating\\
			$\bm{F}^m$ & The unimodal network of modality $m$\\
			$\bm{F}^M$ & The multimodal fusion network\\
			\hline
	\end{tabular}}
\end{table*}%

\subsection{Unimodal Networks}\label{unimodal}
First of all, we illustrate the structure of unimodal networks and the procedures to obtain the unimodal representations for the later fusion and the generation of unimodal labels.
To make a fair comparison with our closest baseline \cite{mmsa}, we use the same structure of unimodal networks as that in Self-MM \cite{mmsa}.
Following the state-of-the-art algorithms \cite{MISA,MAG-BERT,hycon,mmsa,SUGRM}, BERT \cite{BERT} is used to extract the high-level language representation. 
Specifically, the procedures of the BERT network are shown as below: 
\begin{equation}
	\label{eq6}
	\begin{split}
		&\bm{\hat{x}}_{l}=\text{BERT}(\bm{U}_l; \theta_{BERT})\\
		&\bm{x}_{l}=f(\bm{W}_{down}\bm{\hat{x}}_{l} + \bm{b}_{down})\in \mathbb{R}^{d_l\times 1}\\
	\end{split}
\end{equation}
where $\bm{U}_l$ is the sequence of the input tokens, $\bm{\hat{x}}_{l}$ is the representation of the first token of the BERT output sequence. We use a fully connected layer to further process $\bm{\hat{x}}_{l}$ to project the language representation into a low-dimensional feature space. Notably, for a fair comparison with state-of-the-art method \cite{unimse}, we also present the results of MUG with T5 \cite{t5} as the language network. 
For the acoustic and visual modalities, following our baseline \cite{mmsa}, we use the LSTM network as the unimodal network:
\begin{equation}
	\label{eq6}
	\begin{split}
		\bm{x}_{m}=\text{LSTM}(\bm{U}_m; \theta_{m}) \in \mathbb{R}^{d_m\times 1}
	\end{split}
\end{equation}
where $\bm{U}_m\in \mathbb{R}^{T_m \times d'_m}$ is the input feature sequence ($T_m$ denotes the sequence length, $d'_m$ is the input feature dimensionality, and $m \in \{ a, v\}$).
The generated unimodal representation $\bm{x}_{m}$ is used for fusion and learning the unimodal labels.

\subsection{Pre-training of Multimodal Framework}
In this stage, we train the framework to initialize parameters and propose the contrastive-based projection module (CPM) to prepare the framework for the meta-learning stage.  

Formally,  given three unimodal sequences $\bm{U}_m\in \mathbb{R}^{T_m \times d'_m}$ ($m \in \{ l, a, v\}$), a traditional multimodal learning system can be formulated as:
\begin{equation}
	\label{eq111}
	\bm{x}_{m}=\bm{F}^m(\bm{U}_m; \theta_m)\in \mathbb{R}^{d_m\times 1}, \ m \in \{ l, a, v\}
\end{equation}
\begin{equation}
	\label{eq222}
	\bm{x} = \bm{F}^M(\bm{x}_{l}, \bm{x}_{a}, \bm{x}_{v}; \theta_M)\in \mathbb{R}^{d\times 1}
\end{equation}
\begin{equation}
	\label{eqM2}
	\hat{y} = \bm{P}^M(\bm{x}; \theta_{p_M})
\end{equation}
\begin{equation}
	\label{eq333}
	\mathcal{L}_M =\frac{1}{n}\sum_{j=1}^n |y^j-\hat{y}^j|,\ \ \theta \gets \theta - \alpha \nabla_{\theta} \mathcal{L}_M 
\end{equation}
where $\hat{y}$ is the prediction based on multimodal representation $\bm{x}$, $\bm{P}^M$ is the multimodal predictor parameterized by $\theta_{p_M}$, $\bm{F}^m$ is the unimodal network parameterized by $\theta_m$, and $\bm{F}^M$ is the multimodal fusion network  parameterized by $\theta_M$. $y^j$ is the ground truth label for multimodal sample $j$ (the superscribe $j$ is omitted in most equations for brevity), $\theta \in \{\theta_a,\theta_v,\theta_l,\theta_M \}$, $\alpha$ is the learning rate, $n$ is the batch size, and $\mathcal{L}_M$ is the mean absolute error (MAE). For conciseness, the structures of $\bm{F}^m$, $\bm{F}^M$ and $\bm{P}^m$ are illustrated in Fig.~\ref{structure}.

In our framework, to prepare the model for the proposed meta-learning strategy, we first need to train a unimodal predictor for each modality that is able to process both the corresponding unimodal representation and multimodal embedding, such that we can use the multimodal annotations to help learn the unimodal labels for unimodal representations (a more detailed illustration is provided in the next section). To achieve this goal, we design a learnable contrastive-based projection module (CPM) for each modality to project the multimodal representation into unimodal embedding space, and use the projected multimodal embedding to train the unimodal predictor. Thereby, we have the following equations:
\begin{equation}
	\label{equ}
	\setlength{\abovedisplayskip}{1pt}
	\setlength{\belowdisplayskip}{1pt}
	\bm{x}_{m'}=f(\bm{W}_m\bm{x} + \bm{b}_m),\ m \in \{ l, a, v\}
\end{equation}
\begin{equation}
	\label{equ233}
	\setlength{\abovedisplayskip}{0pt}
	\setlength{\belowdisplayskip}{0pt}
	\hat{y}_{m'}=\bm{P}^m(\bm{x}_{m'}; \theta_{p_m})
\end{equation}
\begin{equation}
	\label{eq333}
	\setlength{\abovedisplayskip}{0pt}
	\setlength{\belowdisplayskip}{0pt}
	\mathcal{L}_{m'} =\frac{1}{n}\sum_{j=1}^n |y^j-\hat{y}_{m'}^j|
\end{equation}
where $\bm{W}_m$ and $\bm{b}_m$ are the learnable parameters of the projection layer in the CPM for modality $m$, and $f$ is the activation function. In Eq.~\ref{equ233}, we use the projected multimodal embedding $\bm{x}_{m'}$ to train the unimodal predictor $\bm{P}^m$, such that the unimodal predictor can process multimodal embedding to guide the learning of meta uni-label correction networks for individual modalities at the meta-learning stage.

The projection layer maps the input multimodal representation into a metric space where it can be directly compared with the corresponding modality in terms of similarity. Nevertheless,
projection layer alone is not expressive enough to reduce the distributional gap between unimodal and multimodal embeddings, and contrastive learning \cite{selfcon1, MMIM,lin2022multimodal} is thereby used to further achieve this goal.
Firstly, we perform L2 normalization on the representations:
\begin{equation}
	\setlength{\abovedisplayskip}{3pt}
	\setlength{\belowdisplayskip}{3pt}
	\label{con1}
	(\bm{x}^{*}_{*})_i \longleftarrow \frac{(\bm{x}^{*}_{*})_i}{\sqrt{\sum_{o} (\bm{x}^{*}_{*})_o^2 }}, \ i \in \{1, 2,...\}
\end{equation}
where $*$ represents any one of the possible subscribes or superscribes, $o$ is the index at the feature dimension, and L2 normalization is used to constrain the values of similarity between representations to be -1 to +1.
Then, we use dot product to measure the similarity between anchor and positives/negatives, and calculate the contrastive learning loss as follows:
\begin{equation}
	\setlength{\abovedisplayskip}{1pt}
	\setlength{\belowdisplayskip}{1pt}
	\label{con2}
	\mathcal{L}_{m}^c= -\frac{1}{n}\sum_{j=1}^n log \frac {e^{ \frac{(\bm{x}^j_{m'})^T\bm{x}^j_{m}}{\tau}} } {\sum_{g=1}^{n}e^{\frac{(\bm{x}^j_{m'})^T\bm{x}^g_{m}}{\tau}}}
\end{equation}
where $\tau$ is the temperature parameter. The contrastive learning is used to improve the mutual information between unimodal and multimodal representations and reduce the gap between them. In this way, we can directly use projected multimodal representations and multimodal labels to guide the learning of unimodal labels and estimate the quality of learned unimodal labels. And the unimodal modules can process both unimodal and projected multimodal signals given that they are close in terms of embeddings. Notably, in the unimodal-multimodal contrastive learning, we stop the gradients from the unimodal representations and only change the distributions of the projected multimodal representations to force the projected multimodal representations to have the same distribution as that of the unimodal representations. 

The total loss for the first stage is defined as:
\begin{equation}
	\label{eqstage1}
	\setlength{\abovedisplayskip}{3pt}
	\setlength{\belowdisplayskip}{3pt}
	\mathcal{L} = \mathcal{L}_M + \sum_m (\eta \cdot \mathcal{L}_{m'} + \gamma \cdot\mathcal{L}_{m}^c )
\end{equation}
where $\eta$ and $\gamma$ are the weights of training losses.


\subsection{Meta-Learning of MUCN}
In this section, we learn clean labels of individual modalities based on the weak supervision from multimodal labels via the designed meta-learning strategy. In the meta-training stage, we propose the unimodal denoising task and design the meta uni-label correction network (MUCN) to generate the learned unimodal labels, where MUCN is trained based on the loss of unimodal denoising task. In the meta-testing phrase, we design multimodal denoising task and use multimodal representations (which have sentiment annotations) to evaluate the effectiveness of the updated MUCN. If MUCN becomes less discriminative after meta-training, we compute the meta-gradients of MUCN with respect to the loss of multimodal denoising task, and then meta-update MUCN. In this way, MUCN can be trained more accurately and generate more reasonable unimodal labels for unimodal representations. In contrast to the non-parametric methods \cite{mmsa,SUGRM}, the proposed unimodal learning strategy is more expressive and can estimate the quality of the defined unimodal labels. And compared to other meta-learning strategies \cite{mlc,wu2021learning,noisy_cvpr_2023}, 
our method provides explicit supervision signal for MUCN via the proposed unimodal and multimodal denoising tasks, which is more stable in training and ensures the effective learning of MUCN.

\subsubsection{Unimodal Denoising Task}
Firstly, in the meta-training stage, we train MUCN (the structure of MUCN is shown in Fig.~\ref{structure}) by guiding it to learn to denoise the manually corrupted multimodal label and recover the original multimodal label $y$ using the unimodal representation:
\begin{equation}
	\label{eqstage2}
	\setlength{\abovedisplayskip}{3pt}
	\setlength{\belowdisplayskip}{3pt}
	y_{m_c}\! =\! MUCN_m (\bm{x}_m, y + \epsilon_m; \theta_{m_c}),\ \epsilon_m \!\sim\! \mathcal{N}(0,I)
\end{equation}
\begin{equation}
	\label{eqmeta2233}
	\mathcal{L}_{MUCN}^{j}(\theta_{m_c}) = |y^j - y_{m_c}^j|,\ j \in \{1,2,...,n\}
\end{equation}
\begin{equation}
	\label{eqmeta22}
	\theta'_{m_c} = \theta_{m_c} -\frac{\alpha}{|S|}  \cdot \sum_{j \sim S} \nabla_{\theta_{m_c}} \mathcal{L}_{MUCN}^{j}(\theta_{m_c})
\end{equation}
where the MUCN for modality $m$ ($MUCN_m$) takes as input the unimodal representation and the corrupted multimodal label, and outputs a corrected label $y_{m_c}$ that is expected to match the original multimodal label $y$ by the training using gradient descent. Following MAML \cite{9}, the updated parameters $\theta'_{m_c}$ are computed using one or more gradient descent updates on the unimodal denoising tasks. $S$ denotes the task set that contains a batch of input. The Gaussian noise $\epsilon_m$ prevents $MUCN_m$ to learn identity mapping and provides the model the capability to learn (at least) a sub-optimal unimodal label for the input unimodal representation. 
This training strategy enables a more stable and effective learning of MUCN compared to training MUCN without explicit supervision \cite{mlc}. To be more specific, meta label correction (MLC) \cite{mlc} imposes no additional constraints on the label correction of noisy samples during the meta-training stage, but merely relies on the meta-gradients computed using the clean samples to guide the learning of noisy labels via a bi-level optimization strategy. In contrast to MLC, the proposed strategy can at least learn the corresponding sample-specific global label for each modality via the unimodal denoising task during the meta-training stage. Thereby, our training strategy provides a good initialization for the parameters of MUCN and prevents MUCN from falling into a worse local optimum after meta-training, enabling a stable learning. Our experiment also verifies that the proposed meta-learning strategy outperforms MLC with a considerable margin (see Section ~\ref{sec:ablation}).

\subsubsection{Additional Design for Unimodal Denoising Task}
After the halfway of the meta-learning where $MUCN_m$ becomes discriminative and can generate meaningful labels, we change the objective of the unimodal denoising task (Eq.~\ref{eqmeta2233}) as:
\begin{equation}
	\label{eqmeta2234}
	\setlength{\abovedisplayskip}{3pt}
	\setlength{\belowdisplayskip}{3pt}
	\mathcal{L}_{MUCN}^{j}(\theta_{m_c}) = |y_{mi}^j - y_{m_c}^j|
\end{equation}
where $y_{mi}^j$ is computed as:
\begin{equation}
	\label{eqmeta2235}
	\setlength{\abovedisplayskip}{2pt}
	\setlength{\belowdisplayskip}{2pt}
	y_{mi}^j = \lambda \cdot y_{m'_c}^j + (1-\lambda) \cdot y^j
\end{equation}
where $y_{m'_c}^j$ is the corrected unimodal label of sample $j$ generated by $MUCN_m$ at the previous epoch, and $\lambda$ is a coefficient which is less than 1. In this way, we can leverage the discriminative power of $MUCN_m$ to improve the unimodal denoising task and further prevent the $MUCN_m$ from defining  unimodal label as the corresponding multimodal label. Notably, $\lambda$ decays as the training deepens:
\begin{equation}
	\label{eqmeta2235}
	\setlength{\abovedisplayskip}{3pt}
	\setlength{\belowdisplayskip}{3pt}
	\lambda \longleftarrow (\lambda_{init})^{E+1}
\end{equation}
where $\lambda_{init}$ is the initialized value of $\lambda$, $E$ denotes the current epoch number of meta-learning.

\subsubsection{Multimodal Denoising Task}
Nevertheless, the training strategy of unimodal denoising task still tends to define unimodal label as the corresponding multimodal label. Although multimodal label provides abundant prior information to the unimodal label and they are equal/similar in some cases, this assumption inevitably introduces noise to the training of MUCN as elaborated in the Introduction section. To address this issue, we innovatively design the multimodal denoising task to leverage clean multimodal labels and representations to guide the learning of MUCN. Specifically, we first estimate the effectiveness of $MUCN_m$ by estimating whether it can recover clean multimodal label given the projected multimodal representation and corrupted label:
\begin{equation}
	\label{eqstage23}
	\setlength{\abovedisplayskip}{3pt}
	\setlength{\belowdisplayskip}{3pt}
	\Bar{y}_{c}\! =\! MUCN_m (\bm{x}_{m'}, \Bar{y}; \theta_{m_c})
\end{equation}
\begin{equation}
	\label{eqstage233}
	\mathcal{L}_{MUCN}^{pre}(\theta_{m_c}) = \frac{1}{|D|} \sum_{k \sim D} |y^k - \Bar{y}_{c}^k|
\end{equation}
where $\Bar{y}$ is the corrupted noise label, and $D$ denotes the multimodal denoising task set. To obtain a more accurate estimation of $MUCN_m$ and avoid the bias caused by the same labels of unimodal and multimodal denoising tasks, we additionally randomly sample more data to constitute $D$, i.e., $D=S\bigcup S'$ where $|S'| = b \cdot |S|$ ($b$ is set to 10 in our experiments). Notably, thanks to the CPM at the first stage, the project multimodal representation and corresponding unimodal representation have similar distributions, and therefore it is appropriate to feed them into the same MUCN. 

There is still one question remaining: How to construct the noisy multimodal label $\Bar{y}$? In our setting, we feed the projected multimodal representation $\bm{x}_{m'}$ into the unimodal predictor to obtain the prediction $\hat{y}_{m'}$ (see Eq.~\ref{equ233}) and add a Gaussian noise to the generated prediction, which is used as the noisy multimodal label $\Bar{y}$. This is because the generated prediction $\hat{y}_{m'}$ is semantically related to the true label $y$, and the Gaussian noise is used to increase the difficulty of learning and prevent $MUCN_m$ to learn an inverse mapping of the unimodal predictor. The equation is shown as below:
\begin{equation}
	\label{eqstage24}
	\setlength{\abovedisplayskip}{3pt}
	\setlength{\belowdisplayskip}{3pt}
	\Bar{y} = \hat{y}_{m'} + \epsilon_{m'},\ \epsilon_{m'} \sim \mathcal{N}(0,I)
\end{equation}

\subsubsection{Optimization Strategy}
After the training on unimodal denoising task (see Eq.~\ref{eqmeta22}), $\theta_{m_c}$ is updated to $\theta'_{m_c}$, and we feed the same multimodal input to go through Eqs.~\ref{eqstage23} and ~\ref{eqstage233} again, which generates the post correction loss $\mathcal{L}_{MUCN}^{post}(\theta'_{m_c})$. Then, we estimate the values of $\mathcal{L}_{MUCN}^{post}(\theta'_{m_c})$ and $\mathcal{L}_{MUCN}^{pre}(\theta_{m_c})$. If $\mathcal{L}_{MUCN}^{post}(\theta'_{m_c})$ $ < \mathcal{L}_{MUCN}^{pre}(\theta_{m_c})$ which suggests that $MUCN_m$ becomes more discriminative after meta-training and multimodal label $y$ is likely to be an excellent estimation of unimodal label, we simply skip the meta-testing stage and define the updated parameters of $MUCN_m$ as:
\begin{equation}
	\label{eqstage25}
	\setlength{\abovedisplayskip}{2pt}
	\setlength{\belowdisplayskip}{2pt}
	\theta_{m_c} \longleftarrow \theta'_{m_c}
\end{equation}
And if $\mathcal{L}_{MUCN}^{post}(\theta'_{m_c})$ $ \geq \mathcal{L}_{MUCN}^{pre}(\theta_{m_c})$ which suggests that the corresponding multimodal label $y$ is not an excellent estimation of unimodal label and the corrected label $y_{m_c}$ is actually a more suitable unimodal label, we use a bi-level optimization strategy to correct the `false' update of $MUCN_m$ via meta-updating:
\begin{equation}
	\label{eqmeta26}
	\setlength{\abovedisplayskip}{3pt}
	\setlength{\belowdisplayskip}{3pt}
	\theta_{m_c} \longleftarrow \theta_{m_c} -\alpha'  \cdot \nabla_{\theta_{m_c}} \mathcal{L}_{MUCN}^{post}(\theta'_{m_c})
\end{equation}
where $\alpha'$ is the learning rate at the meta-testing stage. Note that here we directly take the derivative of the original $\theta_{m_c}$ with respect to $\mathcal{L}_{MUCN}^{post}(\theta'_{m_c})$ (that is why it is a bi-level optimization strategy). 
The bi-level optimization strategy is based on the assumption that if the current $MUCN_m$ is discriminative enough, then it is able to perform well on multimodal denoising task whose annotated label is available and features are of similar distribution. The meta-update of $\theta_{m_c}$ encourages $MUCN_m$ to learn to perform well on label denoising task. 
Compared to other meta-learning based methods \cite{mlc,wu2021learning,noisy_cvpr_2023}, the proposed denoising tasks provide explicit supervision for MUCN and our bi-level optimization does not involve the update of other modules, which enables a more stable and effective training of MUCN.

Since only MUCN is updated at this stage, we do not go through the network again but simply use the saved unimodal and multimodal representations generated by the network at the first stage, which can greatly accelerate the training of MUCN.

\subsection{Multi-Task Training}
In this stage, we conduct joint training of unimodal and multimodal learning tasks as done in \cite{mmsa}. We use the corrected unimodal labels generated by MUCN in the meta-learning stage to perform the training of unimodal tasks, and use the multimodal labels to guide the learning of multimodal task. By this means, we can extract more discriminative unimodal features and leverage the available multimodal information to the utmost extent. The training objectives of this stage are:
\begin{equation}
	\label{equ2332}
	\setlength{\abovedisplayskip}{0pt}
	\setlength{\belowdisplayskip}{0pt}
	\hat{y}_{m}^j=\bm{P}^m(\bm{x}_{m}^j; \theta_{p_m}), \ j \in \{1,2,...,n\}
\end{equation}
\begin{equation}
	\setlength{\abovedisplayskip}{0pt}
	\setlength{\belowdisplayskip}{0pt}
	\mathcal{L}_m =\frac{1}{n}\sum_{j=1}^n |y^j_{m_c}-\hat{y}_{m}^j|, \ m \in \{l,a,v\}
\end{equation}
\begin{equation}
	\setlength{\abovedisplayskip}{0pt}
	\setlength{\belowdisplayskip}{0pt}
	\mathcal{L}_M =\frac{1}{n}\sum_{j=1}^n |y^j-\hat{y}^j|
\end{equation}
\begin{equation}
	\label{eqstage1}
	\setlength{\abovedisplayskip}{0pt}
	\setlength{\belowdisplayskip}{0pt}
	\mathcal{L} = \mathcal{L}_M + \sum_m (\beta \cdot \mathcal{L}_m)
\end{equation}
where $\hat{y}_{m}^j$ is the unimodal prediction of sample $j$ generated by unimodal predictor $\bm{P}^m$ based on unimodal representation $\bm{x}_m^j$, and $y^j_{m_c}$ is the corrected label generated by MUCN. 
Note that this stage and the first two stages can be trained separately, that is, we only run previous two stages for one time and get the learned unimodal labels for the third stage, which can greatly reduce the training time. We do not reuse the weights learned at the pre-training stage but train the whole network from scratch.

\section{\textbf{Experiments}}\label{sec:Experiments}
In this section, we conduct extensive experiments to evaluate our MUG using \textbf{CMU-MOSI} \cite{Zadeh2016Multimodal}, \textbf{CMU-MOSEI} \cite{MOSEI}, and  \textbf{SIMS} \cite{sims} datasets.

\subsection{Datasets}

1) \textbf{CMU-MOSI} \cite{Zadeh2016Multimodal} is a prominent resource in the domain of MSA. The CMU-MOSI dataset is collected from online websites, which encompasses a collection exceeding 2,000 video utterances. Sentiment intensity for each video utterance is quantified on a Likert scale from -3 to +3, with the extremes representing the most profound negative and positive sentiments, respectively. 
In alignment with previous research \cite{mmsa,MMIM}, we utilize 1,284 utterances for training purposes, 229 utterances for validation, and 686 utterances for the testing phase.

2) \textbf{CMU-MOSEI} \cite{MOSEI} is a large-scale dataset for MSA, comprising over 22,000 utterances sourced from a diverse group of more than 1,000 YouTube speakers across 250 unique topics.
The selection of utterances spans a wide array of topics and monologues, with each utterance being annotated on two aspects: an emotion label of six distinct emotional states and a sentiment score that is in the range between -3 and +3.
In our MUG, we harness the sentiment labels provided by the CMU-MOSEI dataset to execute the MSA task. 
In congruence with the experimental setups of prior research \cite{mmsa,MMIM}, we have delineated our dataset usage as follows: 16,326 utterances for training, 1,871 utterances for validation, and 4,659 utterances for the testing phase.


3) \textbf{SIMS} \cite{sims} stands out as a unique resource for Mandarin Chinese MSA. It is comprised of 2,281 curated utterances extracted from a variety of sources including films, television series, and variety shows, which capture spontaneous expressions, diverse head poses, occlusions, and varying lighting conditions. Each video utterance is meticulously evaluated by human annotators and assigned a sentiment score ranging from -1, indicating a strongly negative sentiment, to 1, signifying a strongly positive sentiment.
In alignment with the methodologies employed in existing literatures \cite{mmsa,MMIM}, our MUG utilizes following dataset partitioning strategy for model development: 1,368 utterances are designated for training, 456 for validation, and 457 for the testing phase.

\subsection{Evaluation Metrics}


We utilize the following set of evaluation metrics to assess the efficacy of our MUG: 1) Acc7: This metric measures 7-class accuracy, which is the fine-grained categorization of sentiment scores; 2) Acc2: This denotes binary accuracy, distinguishing between positive and negative sentiments; 3) F1 Score: A harmonic mean of precision and recall for binary sentiment classification; 4) MAE: Mean Absolute Error, quantifying the average magnitude of the errors between sentiment predictions and annotated labels; and 5) Corr: The correlation coefficient, which indicates the strength and direction of a linear relationship between the model's predictions and sentiment labels.
It is important to note that for the computation of Acc7, the model's predictions are rounded to the nearest integer within the range from -3 to +3. Additionally, when calculating the Acc2 and the F1 Score, neutral utterances are omitted from consideration.

\subsection{Feature Extraction Details}\label{sec:fea_detail}
\textbf{Visual Modality}:
For the CMU-MOSI and CMU-MOSEI datasets,
Facet$^1$ is employed to extract a sequence of visual features which encompass facial action units, facial landmarks, head pose, and so on.  These visual features are extracted from each utterance at the frequency of 30Hz, creating a temporal sequence of facial expressions. 
In the case of the SIMS dataset, MTCNN face detection algorithm \cite{zhang2016joint} is utilized to extract aligned faces. Subsequently, MultiComp OpenFace2.0 toolkit \cite{baltrusaitis2018openface} is deployed to extract a set of 68 facial landmarks, 17 facial action units, head pose, head orientation, and eye gaze. 

\textbf{Acoustic Modality}:
For the CMU-MOSI and CMU-MOSEI datasets,
COVAREP \cite{Degottex2014COVAREP} is utilized to extract a sequence of acoustic features. The acoustic features include 12 Mel-frequency cepstral coefficients, pitch tracking, speech polarity, glottal closure instants, spectral envelope, etc. They are extracted from the complete audio clip of each utterance at the sampling rate of 100Hz to construct a sequence that captures the variations in the tone of voice throughout the utterance. 
For the SIMS dataset, we employ the LibROSA \cite{mcfee2015librosa} speech toolkit, configured with its default parameters, to acquire acoustic features at a sampling rate of 22050Hz. The extraction process yields 33-dimensional frame-level acoustic features, which include a 1-dimensional logarithmic fundamental frequency (log F0), a set of 20-dimensional Mel-frequency cepstral coefficients (MFCCs), and a 12-dimensional Constant-Q chromatogram (CQT).

\textbf{Language Modality}:
For all the datasets,
following the state-of-the-art methods \cite{MAG-BERT,MISA,CM-BERT,hycon,mmsa}, BERT \cite{BERT} is used to extract the high-level textual representations. More recently, UniMSE \cite{unimse} uses T5 \cite{t5} as the language backbone and obtains much better performance. For a fair comparison with UniMSE \cite{unimse}, we also present the results of our model with T5 \cite{t5} as the language learning network.


For the CMU-MOSEI dataset, the respective input feature dimensions for the language, acoustic, and visual modalities are 768, 74, and 35. In the case of CMU-MOSI, the dimensions for these modalities are 768 for language, 74 for acoustic, and 47 for visual. As for the SIMS dataset, the input dimensions for the three modalities are 768 for language, 33 for acoustic, and 709 for visual.
Note that the features of different modalities are not aligned in our experiment (they have different sequence lengths).

\let\thefootnote\relax\footnotetext{\textsuperscript{\rm 1}iMotions 2017. https://imotions.com/}

\subsection{Experimental Details}\label{sec:exper_detail}

We develop the proposed MUG with the PyTorch framework on RTX2080Ti with torch 1.4.0. The parameters of MUG are updated using AdamW \cite{loshchilov2017decoupled} optimizer. We use the learn2learn package$^2$ to realize the bi-level optimization of meta-learning. To ascertain the values of hyperparameters, we follow previous research \cite{Gkoumas2021WhatMT} to conduct a fifty-times random search to identify the most effective hyperparameter configuration. Subsequently, we train the model with the best hyperparameter setting for five times with different random seeds (1111, 1112, 1113, 1114 and 1115), and we average the testing outcomes of the five-time running to compute the final testing results of the model. This same procedure is employed to derive the results for the baseline models. In the regular training of MUG (the third stage), we perform early-stop where we terminate the training if the validation loss does not decrease for over eight epochs. 
The detailed hyperparameter setting of MUG can be referred to Table~\ref{t2223}. Notably, since we decouple the third stage and the first two stages and only run the first two stages for one time to obtain the corrected unimodal labels, the training of the first two stages does not significantly influence the time complexity of the proposed method. 

\let\thefootnote\relax\footnotetext{\textsuperscript{\rm 2}https://github.com/learnables/learn2learn}

\begin{table}[t]
	\centering
	\caption{ \label{t2223}Hyperparameter Settings of MUG.}
	\resizebox{.95\columnwidth}{!}{\begin{tabular}{c|c|c|c}
			\hline
			& CMU-MOSI & CMU-MOSEI & SIMS \\
			\hline
			Batch Size  &  32 & 16 & 32\\
			Learning Rate  & 3e-5  & 1e-5 & 2e-5 \\
			Pre-Training Epochs &  15 & 10 & 20 \\
			Meta-Training Epochs &  65 & 50 & 80\\
			Meta-training Learning Rate $\alpha$  & 5e-3  & 5e-3  & 5e-3 \\
			Meta-testing Learning Rate $\alpha'$  & 1e-3  &  1e-3 & 1e-3\\
			Contrastive Loss Weight $\gamma$ & 0.01 & 0.01 & 0.01\\
			Projected Predictive Loss Weight $\eta$ & 0.01 & 0.01 & 0.01\\
			Fusion Dimensionality $d$ &  32 & 16 & 128 \\
			Acoustic Dimensionality $d_a$ & 256  & 128 & 256\\
			Visual Dimensionality $d_v$ &  64 & 64 & 256\\
			Language Dimensionality $d_l$ & 64  & 64 & 64 \\
			Unimodal Task Loss Weight $\beta$  & 0.01  & 0.0001  & 0.01\\
			\hline
	\end{tabular}}
	\vspace{-0.1cm}
\end{table}%

\begin{figure*}
	\centering
	\includegraphics[width=0.8\linewidth]{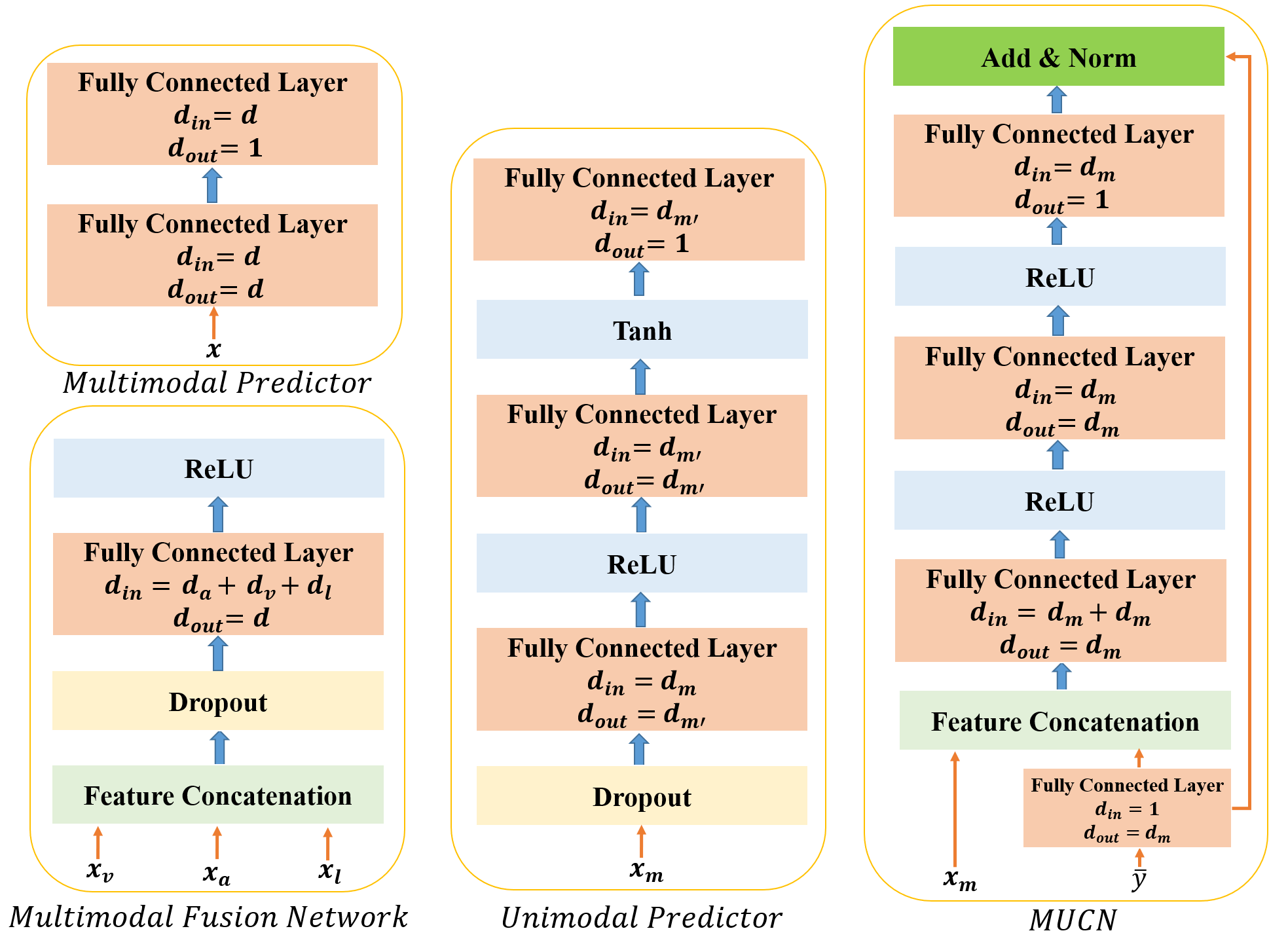}
	\caption{\label{structure}\textbf{The structures of unimodal predictor, multimodal predictor, multimodal fusion network, and MUCN.} We apply a simple fusion network to achieve competitive performance.} 
\end{figure*}

The structures of unimodal predictors, meta uni-label correction networks (MUCNs), multimodal fusion network, and multimodal predictor are shown in Fig.~\ref{structure}. In the final layer of MUCN (i.e., the `Add \& Norm' layer), we conduct residual learning \cite{4} of the input label, and constrain the output value (i.e., the intensity of sentiment) to the desired range which depends on dataset:
\begin{equation}
	\begin{aligned}
		y_{mc} = \rho \cdot tanh(\Bar{y}+y'_{mc})
	\end{aligned}
\end{equation}
where $y'_{mc}$ is the output of the last fully connected layer of MUCN, $\Bar{y}$ is the input noisy label, $y_{mc}$ is the corrected label generated by MUCN, and $\rho$ is the maximum intensity of sentiment that depends on the dataset. In CMU-MOSI and CMU-MOSEI datasets, $\rho$ is equal to 3, and in SIMS dataset, $\rho$ is equal to 1.

\subsection{Baselines}



1) \textbf{Memory Fusion Network} (\textbf{MFN}) \cite{Zadeh2018Memory}, which proposes a delta-attention module and a multi-view gated memory network to capture complicated cross-modal dynamics;
2) \textbf{Multimodal Transformer} (\textbf{MulT}) \cite{MULT}, which generates multimodal representations by translating source modalities into target modalities via  cross-modal Transformers \cite{transformer}; 
3) \textbf{Graph-MFN} \cite{MOSEI}, which designs a multimodal graph neural network to capture unimodal, bimodal, and trimodal interactions and enhance the interpretability of multimodal fusion;
4) \textbf{Tensor Fusion Network} (\textbf{TFN}) \cite{Zadeh2017Tensor}, which leverages outer product over three unimodal features to acquire multimodal tensor, enabling the simultaneous acquisition of unimodal, bimodal, and trimodal interactions;
5) \textbf{Low-rank Modality Fusion} (\textbf{LMF}) \cite{Liu2018Efficient}, which utilizes low-rank weight tensors to address the intricate challenges associated with tensor fusion; 
6)  \textbf{MultiModal InfoMax} (\textbf{MMIM}) \cite{MMIM}, which hierarchically optimizes the mutual information between individual unimodal features and between multimodal representations and unimodal features, thereby enhancing the richness and quality of multimodal representations; 
7) \textbf{All-modalities-in-One BERT} (\textbf{AOBERT}) \cite{aobert}, which introduces a single-stream Transformer that undergoes pre-training on two concurrent tasks: multimodal masked language modeling and alignment prediction, thereby thoroughly exploring the interdependencies and relationships among different modalities; 
8) \textbf{CubeMLP} \cite{cubemlp_mm2022}, which is a multimodal feature processing framework based entirely on MLP that consists of three independent MLP units;
9) \textbf{Hybrid Contrastive Learning} (\textbf{HyCon}) \cite{hycon}, 
 which incorporates both intra-sample and inter-sample contrastive learning mechanisms to thoroughly investigate the interactions within each modality as well as across different modalities;
10) \textbf{Modality-Invariant and -Specific Representations} (\textbf{MISA}) \cite{MISA}, which maps each modality onto two separate subspaces: one that is invariant across modalities and another that is specific to each modality, facilitating a more refined feature fusion;
11) \textbf{Self-Supervised Multi-task
	Multimodal sentiment analysis network} (\textbf{Self-MM}) \cite{mmsa}, which delineates sentiment labels for individual modalities by utilizing the global labels of multimodal samples in a self-supervised fashion, with the aim of extracting more discriminative and expressive features from each unimodal source; 
12) \textbf{Self-supervised Unimodal label Generation using Recalibrated
	Modality representations} (\textbf{SUGRM}) \cite{SUGRM}, which improves Self-MM \cite{mmsa} by projecting unimodal features into a common feature space and enabling a simpler calculation of unimodal labels; 
13) \textbf{Unified MSA and ERC} (\textbf{UniMSE}) \cite{unimse}, which unifies the multimodal sentiment analysis and multimodal emotion recognition tasks and formulates them as generation tasks using auto-regression pre-trained language model.

\begin{table*}
	\centering
	\caption{ \label{tbase}\textbf{The results on CMU-MOSI and CMU-MOSEI datasets.} Except for CubeMLP \cite{cubemlp_mm2022}, AOBERT \cite{aobert} and UniMSE \cite{unimse}, the results of baselines are obtained in our own experiments. The best results are marked in bold and the second best results are marked with underlines. When calculating Acc2 and F1 score, we exclude the neutral utterances.
	}
	\resizebox{1.9\columnwidth}{!}{\begin{tabular}{c|c|c|c|c|c|c|c|c|c|c}
			\hline
			& \multicolumn{5}{c|}{CMU-MOSI} & \multicolumn{5}{c}{CMU-MOSEI}  \\
			\hline
			& Acc7$\uparrow$  & Acc2$\uparrow$ & F1$\uparrow$ & MAE$\downarrow$ & Corr$\uparrow$ & Acc7$\uparrow$  & Acc2$\uparrow$ & F1$\uparrow$ & MAE$\downarrow$ & Corr$\uparrow$\\
			\hline
			Graph-MFN \cite{MOSEI} & 34.4 &   80.2  &   80.1  &  0.939  &  0.656  &  
			51.9  &  84.0  &   83.8  & 0.569  & 0.725  \\
			MFM \cite{MFM} & 33.3 & 80.0  &   80.1 & 0.948 & 0.664  &  
			50.8 &  83.4  &   83.4  & 0.580 & 0.722 \\
			MISA \cite{MISA} & 43.5 &   83.5  &   83.5  & 0.752 & 0.784 & 
			52.2 &  84.3  &   84.3  & 0.550  & 0.758 \\
			MMIM \cite{MMIM} & 45.0 &   85.1  &   85.0  & 0.738  & 0.781  & 
			53.1 &   85.1  &   85.0  & 0.547  &  0.752\\
			Self-MM \cite{mmsa} &  45.8 &   84.9  &   84.8  & 0.731  & 0.785  & 
			53.0 &   85.2  &   85.2  &  0.540 & 0.763 \\
			HyCon \cite{hycon} & 46.6 &   85.2  &   85.1  &  0.741 & 0.779  & 
			52.8 &    85.4  &   85.6  &  0.554 & 0.751 \\
			SUGRM \cite{SUGRM} & 44.9 &  84.6 &  84.6  &  0.739 &  0.772 & 53.7
			&   85.4  &  85.3  &  0.537 &  0.759 \\
			\hline
			CubeMLP \cite{cubemlp_mm2022} & 45.5 &   85.6  &   85.5  & 0.770  & 0.767  & 
			\textbf{54.9} &   85.1  &   84.5  & 0.529  &  0.760\\
			AOBERT \cite{aobert} & 40.2 &   85.6  &   \underline{86.4}  & 0.856  &  0.700 & 54.4
			&   \underline{86.2}  &   85.9 &  \textbf{0.515} &  0.763 \\
			UniMSE (T5)  \cite{unimse} & \underline{48.7} & \underline{86.9} & \underline{86.4} & \underline{0.691} & \underline{0.809} & 54.4 &  \textbf{87.5} & \underline{87.5} & 0.523 & \underline{0.773} \\
			\hline
			MUG (BERT)  & 47.5 &   85.8 &   85.6  &  0.711 &  0.795 &  54.7 &   \underline{86.2}  & 86.1   &  0.527 &  0.772
			\\
			MUG (T5)  & \textbf{49.1}  &   \textbf{87.5} &  \textbf{87.5}   &  \textbf{0.682} & \textbf{0.816}  & \underline{54.8}  &  \textbf{87.5}   &   \textbf{87.6} &  \underline{0.521} &  \textbf{0.776} \\
			\hline
	\end{tabular}}
\vspace{-0.2cm}
\end{table*}%

\begin{table}[t]
	\centering
	\caption{ \label{tsims}\textbf{ The comparison with baselines on SIMS dataset.} The results of the baselines are obtained using the codes in \url{https://github.com/thuiar/MMSA} and \url{https://github.com/yewon95/SUGRM}.
	}
	\resizebox{.95\columnwidth}{!}{\begin{tabular}{c|c|c|c|c}
			\hline
			& MAE $\downarrow$ & Corr $\uparrow$ & Acc-2 $\uparrow$ & F1-score $\uparrow$ \\
			\hline
			TFN \cite{Zadeh2017Tensor} & 0.432 & 0.591 & 78.38 &  78.62  \\
			LMF \cite{Liu2018Efficient} & 0.441  & 0.576 & 77.77 &  77.88 \\
			
			MFN \cite{Zadeh2018Memory} & 0.435  & 0.582 & 77.90 &  77.88 \\
			Graph-MFN \cite{MOSEI} & 0.445  & 0.578 & 78.77 &  78.21 \\
			MulT \cite{MULT} & 0.453  & 0.564 & 78.56 &  79.66 \\
			Self-MM \cite{mmsa} & 0.425 & 0.592 & 80.04 & \underline{80.44}   \\
			SUGRM \cite{SUGRM} & 0.418 & \underline{0.596} & 79.26 & 79.13   \\
			\hline
			MUG (Without Unimodal Tasks)  &  0.434 & 0.573  &  78.16 & 78.10  \\
			MUG (With True Label)  & \textbf{0.410}  &  \textbf{0.601} & \textbf{80.74}  &  \textbf{80.67} \\
			MUG  &  \underline{0.415} &  \textbf{0.601} & \underline{80.31}  & 80.36  \\
			\hline
	\end{tabular}}
\end{table}%

\subsection{Comparison with Baselines}

In this section, we compare MUG with competitive baselines on MSA. Please note that to make a fair comparison with our closest baseline Self-MM \cite{mmsa}, we adopt simple fusion approach and unimodal networks. 
As shown in Table~\ref{tbase}, using BERT as the language network, the contrastive learning based algorithm HyCon \cite{hycon}, MLP-based method CubeMLP \cite{cubemlp_mm2022}, and pre-training method AOBERT \cite{aobert} obtain very competitive results. Nevertheless,  MUG (BERT) still outperforms them and obtains the best performance on the majority of the evaluation metrics on CMU-MOSI and CMU-MOSEI datasets. Specifically, on CMU-MOSI dataset, MUG (BERT) outperforms HyCon by 0.9\% in Acc7, 0.6\% in Acc2 and 0.5\% in F1 score, and it also outperforms AOBERT by over 7\% in Acc7. The improvement is remarkable considering that the human performance is also at this level \cite{Zadeh2017Tensor}. On CMU-MOSEI dataset, MUG (BERT) yields 0.5\% improvement in Acc2 and 1.1\% improvement in F1 score compared to the competitive baseline CubeMLP, and outperforms the best performing model AOBERT in terms of Acc7, F1 score and Corr. MUG (BERT) also reaches state-of-the-art performance in Acc2, F1 score, and Corr on CMU-MOSEI dataset. Compared to these competitive baselines, MUG explicitly meta-learns accurate unimodal labels and conducts unimodal learning tasks under the supervision of learned unimodal labels, which can extract more expressive and discriminative unimodal features for the multimodal inference. 

Compared to Self-MM \cite{mmsa} that also aims to obtain unimodal labels, MUG (BERT) outperforms it on both datasets with a remarkable margin. Similar results are observed on the SIMS dataset in Table~\ref{tsims}, where MUG outperforms Self-MM in terms of MAE, Corr and Acc-2. Given that our structures of unimodal networks and the multi-task training at the third stage are the same as that of Self-MM, the improvement of MUG over Self-MM indicates the superiority of the proposed unimodal label learning strategy. Moreover, our method also outperforms SUGRM \cite{SUGRM} (an improved version of Self-MM) on all the three datasets with considerable margin.
The results are reasonable because compared to the non-parametric approaches adopted in Self-MM and SUGRM, we meta-learn unimodal labels using the proposed unimodal and multimodal denoising tasks, which is more expressive and can estimate the quality of the learned unimodal labels.
These results demonstrate the effectiveness of MUG, indicating the importance of learning unimodal labels for extracting more discriminative and meaningful unimodal features.

In addition, it is shown in UniMSE \cite{unimse} that using T5 \cite{t5} as the language network outperforms previous state-of-the-art methods that apply BERT \cite{BERT}.
Therefore, for a more comprehensive comparison,
we also present the results of our MUG using T5 as the language network. As presented in Table~\ref{tbase}, using T5 brings consistent improvement to the overall performance of the model on two datasets. Moreover, the MUG (T5) still outperforms current state-of-the-art method UniMSE \cite{unimse}, which demonstrates the superiority of our unimodal label learning strategy. Generally, these results suggest that using more advanced language network brings considerable improvement to the multimodal sentiment analysis systems, and our method consistently outperforms baselines across different language networks, indicating the robustness of the proposed algorithm.

\subsection{The Effectiveness of Annotated Unimodal Labels}

Additionally, to justify that using ground-truth unimodal labels to learn unimodal representations can benefit the performance of multimodal system, we carry out an experiment that uses the annotated unimodal labels to train unimodal networks for the SIMS dataset (the SIMS dataset is the only one that has fine-grained annotations for unimodal signals among the three datasets). The results are shown in the Table~\ref{tsims}. From the results of `MUG (With True Label)' in Table~\ref{tsims}, we can infer that using the annotated unimodal labels to learn unimodal representations improves the overall performance of the model in terms of all the evaluation metrics, especially compared to the MUG version that does not incorporate unimodal learning tasks (see the case of `MUG (Without Unimodal Tasks)' in Table~\ref{tsims}). Moreover, the MUG version that uses the learned unimodal labels to conduct unimodal learning tasks outperforms the MUG version that does not incorporate unimodal learning tasks, and is slightly inferior to the MUG version that uses the ground-truth unimodal labels. The results indicate that the learned unimodal labels are effective but still contain noise.

\subsection{Ablation and Comparison Experiments} \label{sec:ablation}
In this section, we conduct extensive ablation studies and comparison experiments to evaluate the effectiveness of the components in MUG.

\textbf{1) Unimodal Learning Tasks}:
In the case of `W/O Unimodal Tasks' (see Table~\ref{t3}), we remove the unimodal learning tasks and the accompanied meta-learning strategy.
As shown in  Table~\ref{t3}, the performance drops dramatically by about 2 points in Acc7, Acc2 and F1 score, demonstrating the importance of our multi-task training for extracting more discriminative unimodal features and improving the performance of the multimodal system;

\textbf{2) Multimodal Denoising Task}:
In the case of `W/O Multimodal Denoising Task', we remove the multimodal denoising task such that only the unimodal denoising task is retained for training the MUCN. The results suggest that the multimodal denoising task brings about 1\% improvement  to the multimodal system in terms of Acc7, Acc2 and F1 score. 
These results reveal that using multimodal denoising task to meta-learn MUCN is of great significance to unimodal label learning, which can use the information of clean multimodal labels to guide the learning of MUCN;

\textbf{3) Unimodal-Multimodal Contrastive Learning}:
In `W/O Contrastive Learning', the unimodal-multimodal contrastive learning in the CPM that aims to reducing the gap between unimodal and multimodal representations is removed. The results on the majority of the evaluation metrics decrease considerably. We argue that this is because without the contrastive learning, the distribution gap of unimodal and multimodal representations cannot be reduced, and we cannot effectively leverage the available multimodal labels for learning more accurate unimodal labels;

\textbf{4) Defining multimodal label as unimodal label}:
In the case of `$y_{m_c} \longleftarrow y$', we directly assign the multimodal label as the unimodal label, and then jointly train unimodal and multimodal tasks.
We observe that Acc7 and Acc2 drop by about 1\% and the performance on other evaluation metrics also decreases, demonstrating that the meta-learned unimodal labels are better than directly using multimodal labels as unimodal labels;

\textbf{5) Meta-Learning Strategy}:
In the case of `MLC', we replace the proposed meta-learning strategy with another strategy proposed in MLC \cite{mlc}. MLC learns the label of noisy samples given clean samples. However, the training of its label correction network (LCN) does not have explicit supervision but merely relies on the main task loss of clean samples. Moreover, it needs to first update the main net using the corrected labels and then computes the meta-gradients of LCN. The long bi-level backpropagation of MLC might weaken the effect of meta-gradients for updating LCN and cause gradient vanishing, and the simultaneous optimization over both main net and LCN is difficult to achieve an optimal routine \cite{noisy_cvpr_2023}. When MLC is applied,
the performance on all the evaluation metrics declines, suggesting that the proposed meta-learning strategy is better than MLC in terms of learning accurate unimodal labels. This is reasonable because we design unimodal and multimodal denoising tasks to train MUCN with explicit supervision and our strategy does not involve the update of other modules, which is more stable and effective in training.

\begin{table}[t]
	\centering
	\caption{ \label{t3}\textbf{ Ablation and comparison experiments on the CMU-MOSI dataset.} In the case of `MLC', we replace the proposed meta-learning strategy with the meta label correction strategy.
	}
	\resizebox{.95\columnwidth}{!}{\begin{tabular}{c|c|c|c|c|c}
			\hline
			& Acc7 & Acc2 & F1 & MAE & Corr \\
			\hline
			W/O Unimodal Tasks & 45.9 & 83.6 & 83.8 & 0.719 & 0.786\\
			W/O Multimodal Denoising Task & 46.8 & 84.7 & 84.7 & 0.715 & \textbf{0.797} \\
			W/O Contrastive Learning & 46.6 & 85.2 & 85.1 & 0.715 & 0.795 \\
			$y_{m_c} \longleftarrow y$  & 46.6 & 84.8 & 84.8 & 0.718 & 0.794 \\
			MLC \cite{mlc} & 47.1 & 84.9 & 84.9 & 0.713 & 0.794 \\
			\hline
			MUG  & \textbf{47.5} &   \textbf{85.8} &   \textbf{85.6}  &  \textbf{0.711} &  0.795 \\
			\hline
	\end{tabular}}
	\vspace{-0.2cm}
\end{table}%

\begin{figure}[htbp]
	\subfigure[W/O CL]{
		\begin{minipage}[t]{0.45\linewidth}
			\centering
			\includegraphics[width=1.5in]{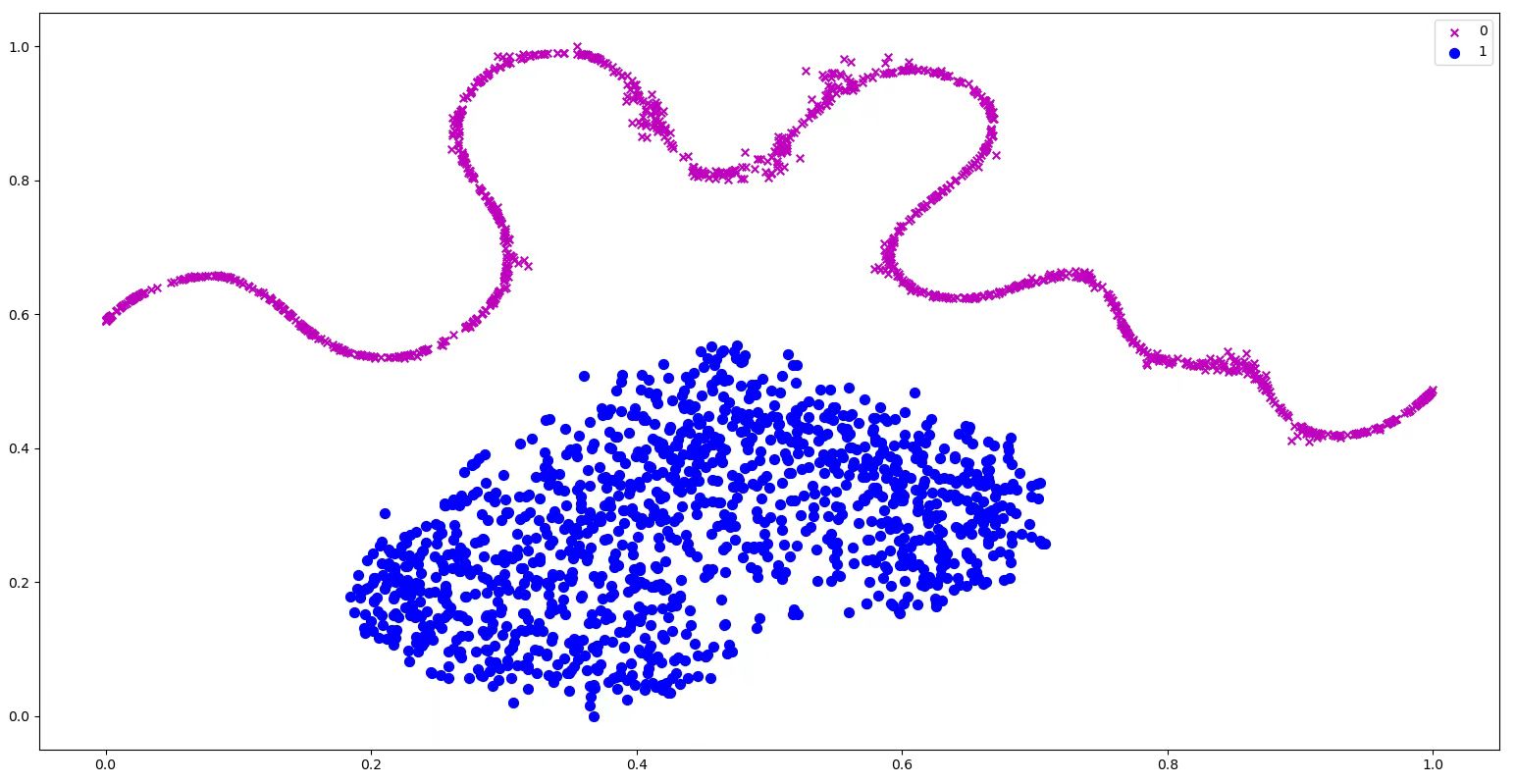}
		\end{minipage}
	}%
	\centering
	\centering
	\subfigure[With CL]{
		\begin{minipage}[t]{0.45\linewidth}
			\centering
			\includegraphics[width=1.5in]{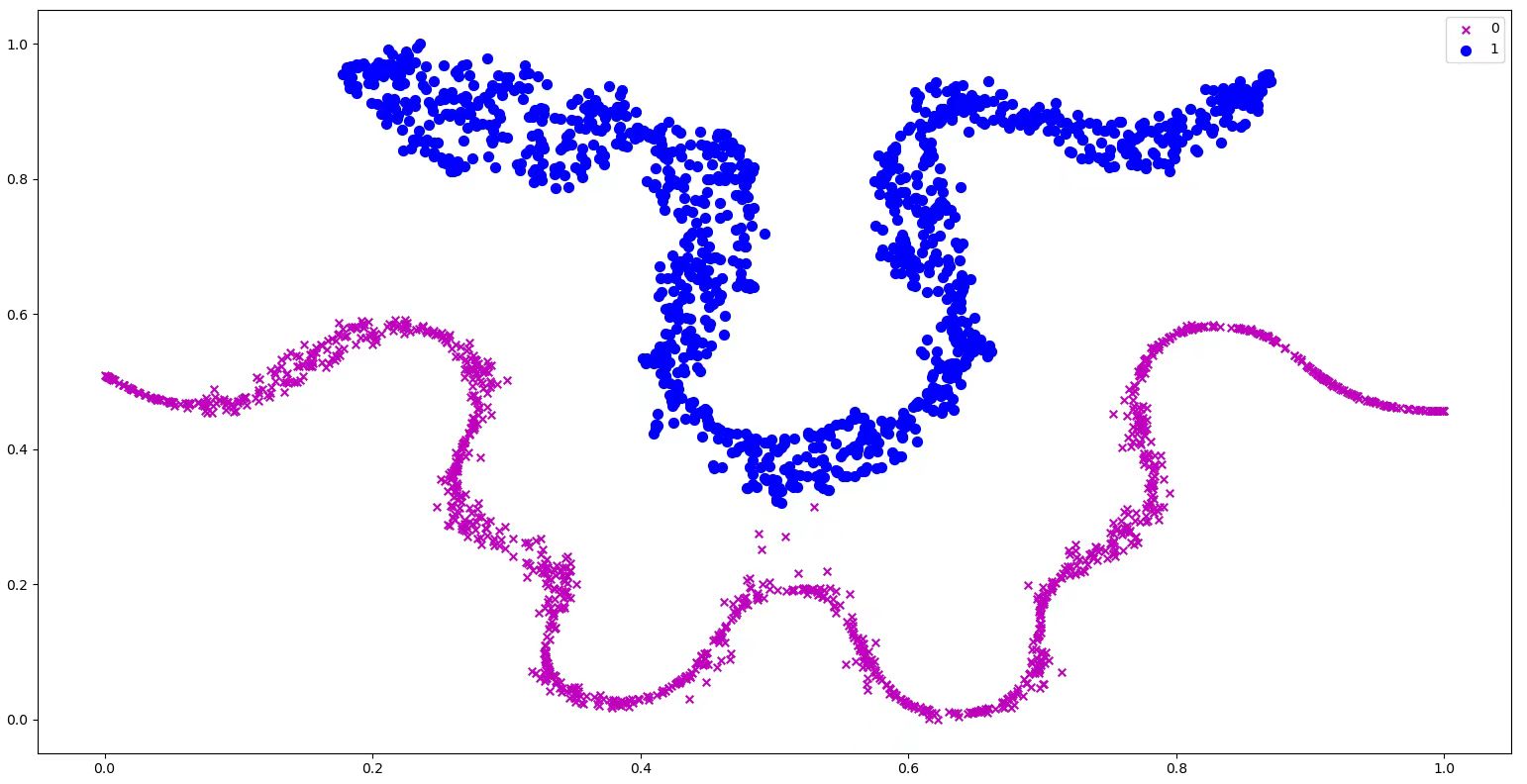}
		\end{minipage}%
	}
	\caption{\label{8} \textbf{T-SNE visualization of unimodal representations and projected multimodal representations.} `CL' denotes contrastive learning. We  use language modality to illustrate the effect of CL, and similar results are observed in other modalities. } 
	\vspace{-0.2cm}
\end{figure}

\subsection{Visualization of Embedding Space}

We provide the visualization for distributions of the projected multimodal representations and the corresponding unimodal representations in the embedding space to analyze the effectiveness of the contrastive-based projection module. 
We perform the visualization experiment by transforming each representation into two-dimensional feature point using the t-SNE \cite{tsne}, which can capture the local structure of the high-dimensional features. 
The visualization experiment is conducted using t-SNE \cite{tsne}. The results are illustrated in Figure~\ref{8}, showcasing the visualizations of language and projected multimodal representations learned with and without contrastive learning.
We can observe that when the contrastive learning is removed, the data points of the projected multimodal representations (depicted in blue) tend to form an oval shape in the embedding space, while the data points of unimodal representations exhibit a striped pattern. In contrast, when the contrastive learning is applied, both the data points from the two distinct sources exhibit a striped pattern, and the gap between them is diminished.
Despite the inevitable existence of distribution gap due to the heterogeneous nature of multimodal and unimodal representations, MUG is able to effectively reduce the gap between them. And their distributional shapes in the embedding space become highly similar, enabling the unimodal modules to process the projected multimodal representations to guide the learning of unimodal labels.

Notably, in the unimodal-multimodal contrastive learning, we stop the gradients from the unimodal representations  and only change the distributions of the projected multimodal representations to force the projected multimodal representations to have the same distribution as that of the unimodal representations. Therefore, the distribution of unimodal representations is almost the same no matter the contrastive learning is applied or not, while the distribution of the projected multimodal representations becomes much similar to the distribution of unimodal representations when the contrastive learning is applied.

\begin{figure*}
	\centering  
	\subfigbottomskip=2pt 
	\subfigcapskip=-5pt 
	\subfigure[Meta-training learning rate $\alpha$]{
		\includegraphics[width=0.48\linewidth]{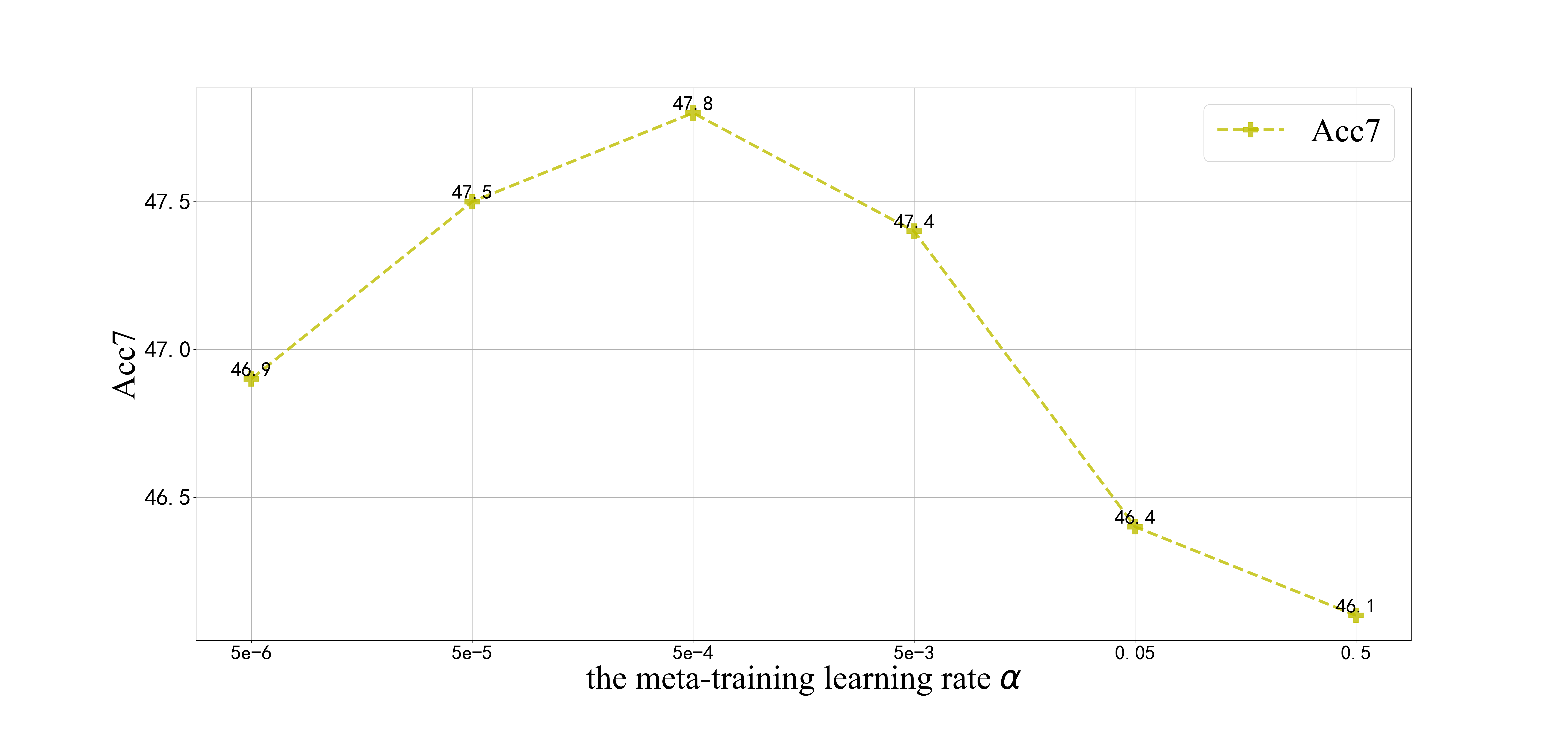}}
	\subfigure[Weight of unimodal learning loss $\beta$]{
		\includegraphics[width=0.48\linewidth]{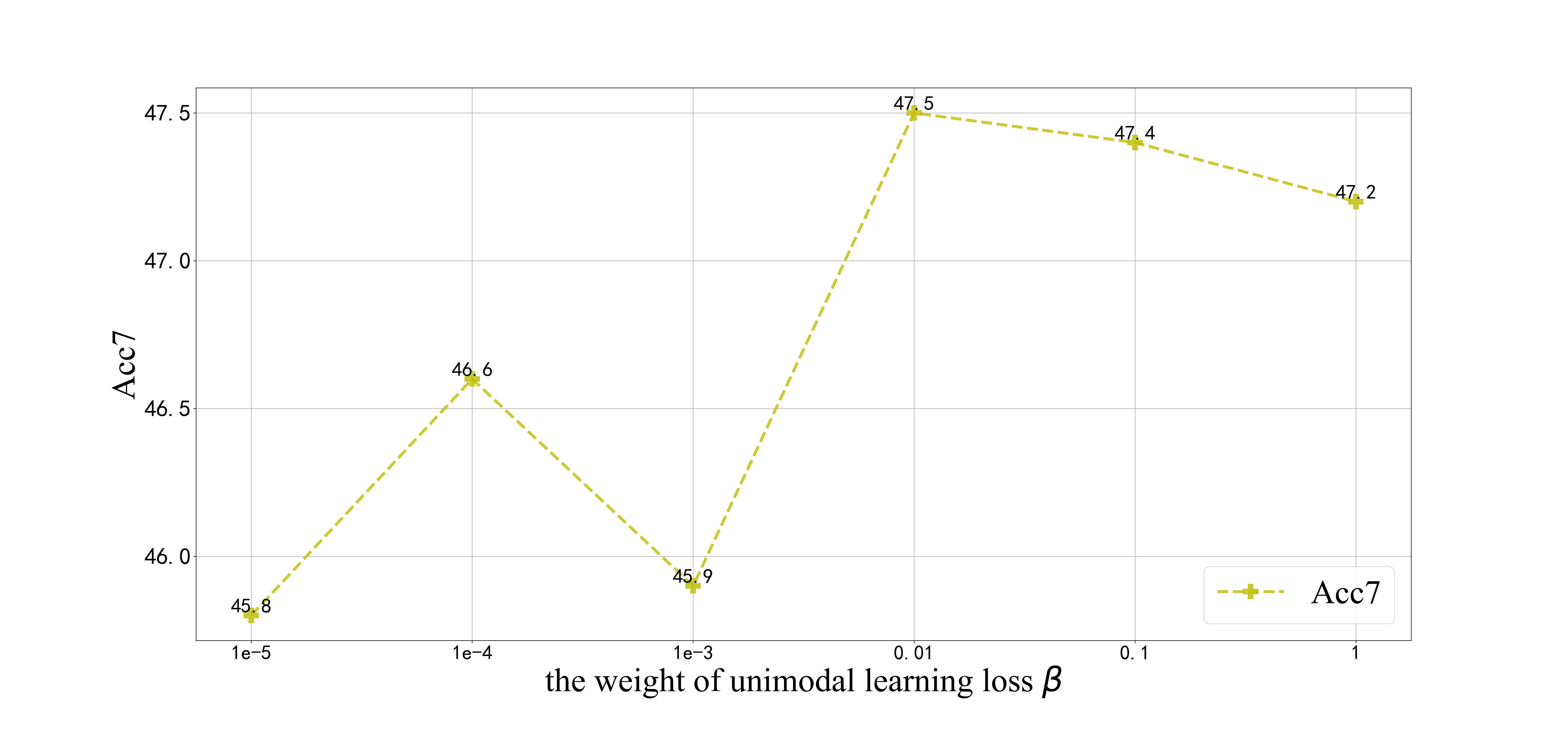}}
	\\
	\caption{\label{hy}\textbf{Model performance with respect to the change of  hyperparameters $\alpha$ and $\beta$.}  }
	\vspace{-0.2cm}
\end{figure*}

\subsection{\textbf{Hyperparameter Robustness Analysis}}

In this section, we evaluate the effects of the hyperparameters $\beta$ (the weight of unimodal learning losses on the third stage) and $\alpha$ (the meta-training learning rate) on CMU-MOSI dataset. The results with $\beta$ and $\alpha$ set to different values are presented in the Fig.~\ref{hy}.

As we can infer from Fig.~\ref{hy} (b), when $\beta$ is set to a relatively large value (0.01-1), the model achieves satisfactory performance. And when $\beta$ is a small value, the performance of the model drops significantly. This is because when $\beta$ is not large enough, the effect of unimodal learning tasks is attenuated, and we cannot extract expressive unimodal features for multimodal learning. As for $\alpha$, the model reaches good performance when $\alpha$ is a moderate value, and its performance declines significantly when $\alpha$ becomes large. This is reasonable because when the learning rate of meta-training becomes large, the update of the parameters in MUCN will become unsteady, and thus we cannot learn effective MUCN to obtain excellent unimodal labels.
Generally, the model can reach satisfactory results over a wide range of hyperparameter settings for $\beta$ and $\alpha$, which to some extent suggests the robustness of the proposed algorithm.

\begin{figure*}[h]
	\centering
	\includegraphics[width=0.92\linewidth]{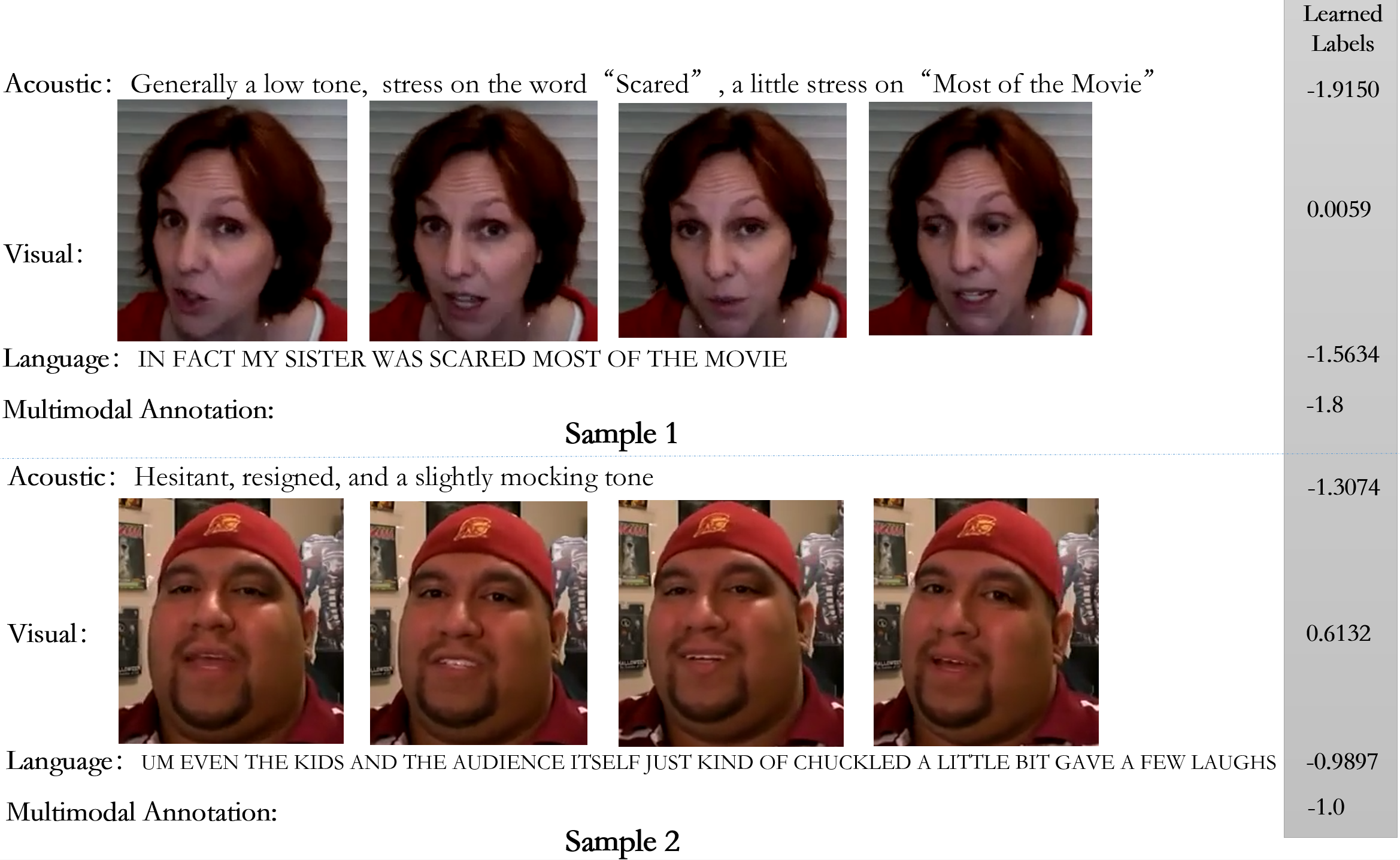}
	\caption{\label{case}\textbf{Case analysis of learned unimodal labels.} }
	\vspace{-0.2cm}
\end{figure*}

\subsection{Case Analysis of Unimodal Labels}

In this section, we provide a case study that visualizes two qualitative samples from CMU-MOSI dataset showing the multimodal input, learned unimodal labels and the multimodal label. As illustrated in Fig.~\ref{case}, for sample 1, the speaker has no obvious facial expression to suggest her emotional tendency, and our method accurately defines the visual label as 0.0059 (almost neutral). As for the language, the speaker describes the film as `scary', and thus the proposed method assigns the language label as -1.5634. For acoustic modality, the speaker stresses on the word `Scared' with a high tone and also stresses on `Most of the Movie', and the model infers the acoustic label as strong negative (-1.9150). It can be inferred that our method can identify the discriminative level of modalities and assigns appropriate labels to them, and specifically can identify non-informative modalities. 

Sample 2 is more difficult to identify because of the existence of contradictory modalities. For visual modality, the speaker has the tendency to laugh at some frames, and therefore the model defines its label as weak positive (0.6132). For language modality, the speaker indicates that the audience laugh at the movie which however does not have obvious sentiment tendency (laugh at the movie can be positive or negative), and the model assigns the language label as -0.9897 (weak negative). However, the speaker uses a resigned and slightly mocking tone which indicates the voice is negative, and the model assigns the acoustic label as -1.3074 (strong negative). The results suggest that MUG can handle the situations where contradictory modalities exist, and by accurately learning unimodal labels, we can extract more discriminative features for accurate multimodal inference.

\subsection{Model Complexity Analysis}

\begin{table}[t]
	\centering
	\caption{ \label{t_com}\textbf{ Comparison on number of parameters.} 
	}
	\begin{tabular}{c|c}
		\hline
		Model & Number of Parameters \\
		\hline
		Self-MM \cite{mmsa} & \textbf{109,647,908} \\
		MISA \cite{MISA} &  110,917,345 \\
		MMIM \cite{MMIM} & 109,821,129 \\
		\hline
		MUG  & 110,089,956 \\
		\hline
	\end{tabular}
	\vspace{-0.2cm}
\end{table}%

\textbf{Space Complexity}: We use the number of trainable parameters as the proxy of space complexity. For the number of parameters, as shown in Table~\ref{t_com}, the proposed framework has a total number of 110,089,956 parameters. In contrast, the number of parameters for MISA and MAG-BERT is 110,917,345. And the number of parameters for Self-MM and MMIM are 109,647,908 and 109,821,129, respectively. Therefore, our model has moderate number of parameters. This is reasonable because we design additional meta uni-label correction network and contrastive-based projection module for each modality. As contrast, Self-MM does not design additional learnable module for learning the unimodal labels. Therefore, our model has slightly more parameters than Self-MM.

\textbf{Time Complexity}: For the training time, as we have stated in the Algorithm section, we only run previous two stages for one time to generate the unimodal labels for the joint training stage (the third stage). The joint training stage and the first two stages can be decoupled, which can greatly reduce the training time. Therefore, the only additional time cost for our MUG is the first two stages that only run for one time and do not need any tuning, which is highly acceptable. As for the third stage (multi-task training stage), since we do not design complex fusion mechanism and do not have to learn unimodal labels at the third stage, the training time of the third stage is faster than our baseline Self-MM which has to calculate unimodal labels during multi-task training. Specifically, under the same environment and same batch size, Self-MM takes about 1.16s to finish one iteration, and our MUG takes about 0.86s to finish one iteration.

\section{Conclusion and Future Work}\label{sec:Conclusion}
In this paper, we focus on the learning of unimodal labels with weak supervision. We propose a novel meta-learning framework to learn the label for each modality based on the available multimodal label, in which we propose unimodal and multimodal denoising tasks to train MUCN with explicit supervision. The multimodal denoising task with clean labels can guide the learning of unimodal denoising task via a bi-level optimization strategy. Then, we jointly train unimodal and multimodal tasks to extract discriminative unimodal features for multimodal inference. Experimental results indicate that MUG outperforms very competitive baselines on MSA. Notably, MUG can be extended to other noisy label learning scenarios. In the future, we aim to evaluate the effectiveness of MUG on various downstream tasks. Moreover, we intend to adapt the proposed meta-learning strategy to handle the label ambiguity problem where the annotated labels of some multimodal samples might be inaccurate due to the subjective and ambiguous nature of sentiment.


%

\appendices

\ifCLASSOPTIONcaptionsoff
  \newpage
\fi



%




\bibliographystyle{IEEEtran}
%
\bibliography{sentiment2}
\end{document}